\def\year{2021}\relax
\documentclass[letterpaper]{article} 
\usepackage{aaai21}  
\usepackage{times}  
\usepackage{helvet} 
\usepackage{courier}  
\usepackage[hyphens]{url}  
\usepackage{graphicx} 
\usepackage{natbib}  
\usepackage{caption} 
\usepackage{amssymb}
\usepackage{amsmath}
\usepackage{verbatim}
\usepackage{color}
\usepackage{multirow}
\usepackage{multicol}
\usepackage[switch]{lineno}  %
\newcommand{\eg}{\emph{e.g.}}
\newcommand{\ie}{\emph{i.e.}}

\title{Write-a-speaker: Text-based Emotional and Rhythmic Talking-head Generation}
\author{
    Lincheng Li,\textsuperscript{\rm 1}\footnotemark[1]
    Suzhen Wang,\textsuperscript{\rm 1}\footnotemark[1]
    Zhimeng Zhang,\textsuperscript{\rm 1}\footnotemark[1]
    Yu Ding,\textsuperscript{\rm 1}\thanks{Equal contribution. Yu Ding is the corresponding author.} \\
    Yixing Zheng,\textsuperscript{\rm 1}
    Xin Yu,\textsuperscript{\rm 2} 
    Changjie Fan\textsuperscript{\rm 1} \\ 
}
\affiliations{
    \textsuperscript{\rm 1} Netease Fuxi AI Lab \\
    \textsuperscript{\rm 2} University of Technology Sydney \\
    \{lilincheng, wangsuzhen, zhangzhimeng, dingyu01, zhengyixing01, fanchangjie\}@corp.netease.com \\
    xin.yu@uts.edu.au 
}

\begin{document}

\maketitle

\begin{abstract}
In this paper, we propose a novel text-based talking-head video generation framework that synthesizes high-fidelity facial expressions and head motions in accordance with contextual sentiments as well as speech rhythm and pauses. To be specific, our framework consists of a speaker-independent stage and a speaker-specific stage. In the speaker-independent stage, we design three parallel networks to generate animation parameters of the mouth, upper face, and head from texts, separately. In the speaker-specific stage, we present a 3D face model guided attention network to synthesize videos tailored for different individuals. It takes the animation parameters as input and exploits an attention mask to manipulate facial expression changes for the input individuals. Furthermore, to better establish authentic correspondences between visual motions (i.e., facial expression changes and head movements) and audios, we leverage a high-accuracy motion capture dataset instead of relying on long videos of specific individuals. After attaining the visual and audio correspondences, we can effectively train our network in an end-to-end fashion. Extensive experiments on qualitative and quantitative results demonstrate that our algorithm achieves high-quality photo-realistic talking-head videos including various facial expressions and head motions according to speech rhythms and outperforms the state-of-the-art.
\end{abstract}

\section{Introduction}

Talking-head synthesis technology aims to generate a talking video of a specific speaker with authentic facial animations from an input speech. 
The output talking-head video has been employed in many applications, such as intelligent assistance, human-computer interaction, virtual reality, and computer games.
Due to its wide applications, talking-head synthesis has attracted a great amount of attention.

Many previous works that take audios as input mainly focus on synchronizing lower facial parts (\eg, mouths), but often neglect animations of the head and upper facial parts (\eg, eyes and eyebrows).
However, holistic facial expressions and head motions are also viewed as critical channels to deliver communicative information \cite{ekman1997face}. For example, humans unconsciously use facial expressions and head movements to express their emotions \cite{Mignault2003}. Thus, generating holistic facial expressions and head motions will lead to more convincing person-talking videos.

Furthermore, since the timbre gap between different individuals may lead the acoustic features in the testing utterances to lying outside the distribution of the training acoustic features, prior arts built upon the direct association between audio and visual modalities may also fail to generalize to new speakers' audios \cite{taida}.
Consequently, the acoustic feature-based frameworks do not work well on input speeches from different people with distinct timbres or synthetic speeches \cite{sadoughi2016head}.

\begin{figure}
\centering
\includegraphics[width=0.45\textwidth]{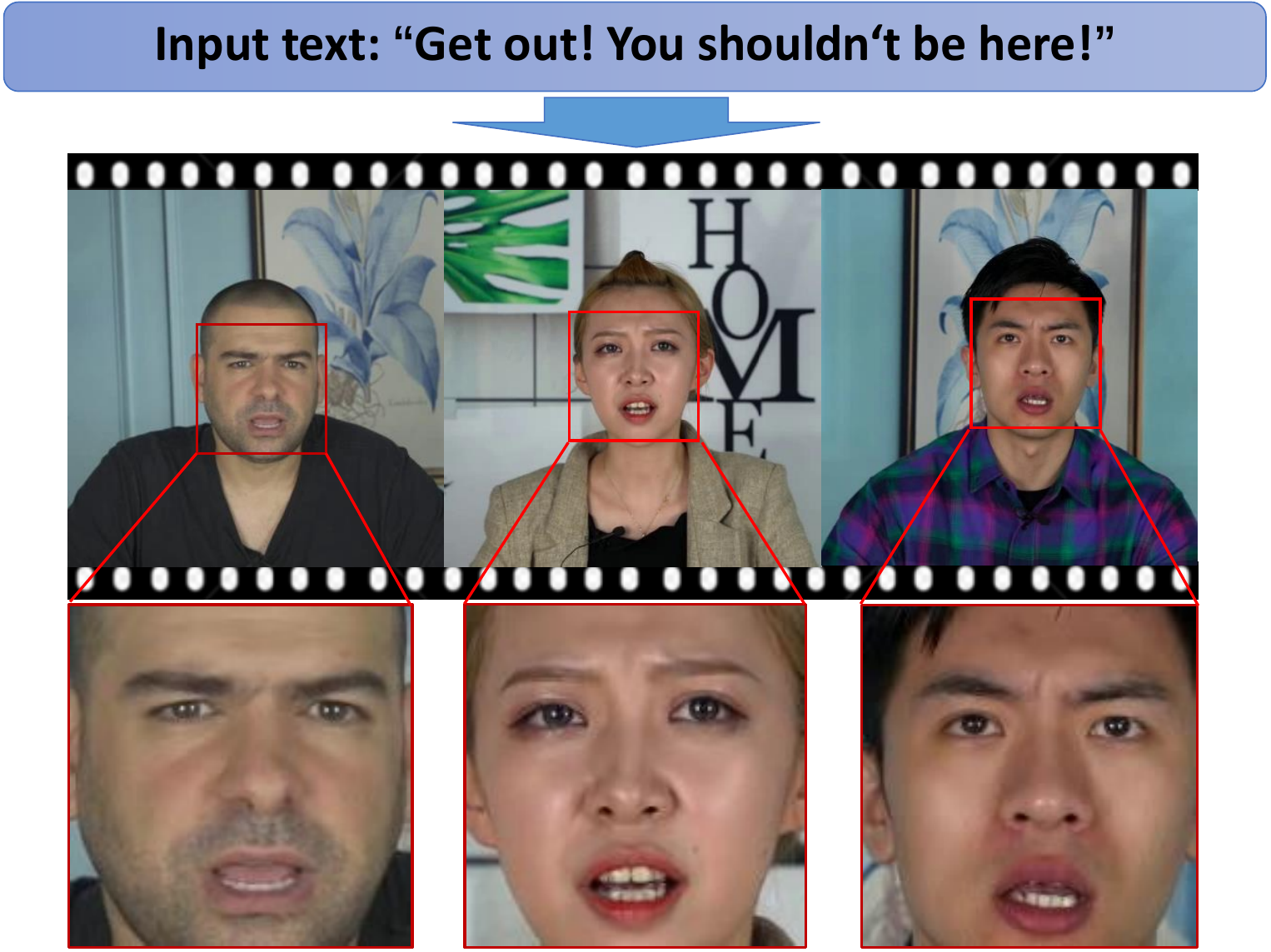}
\caption{Our method produces emotional, rhythmic and photo-realistic talking-head videos from input texts.}
\label{fig:teaser}
\end{figure}

Unlike previous works, we employ time-aligned texts (\ie, text with aligned phoneme timestamps) as input features instead of acoustics features to alleviate the timbre gap issue.
In general, time-aligned texts can be extracted from audios by speech recognition tools or generated by text-to-speech tools. 
Since the spoken scripts are invariant to different individuals, our text-based framework is able to achieve robust performance against different speakers.

This paper presents a novel framework to generate holistic facial expressions and corresponding head animations according to spoken scripts.
Our framework is composed of two stages, \ie, a speaker-independent stage and a speaker-specific stage. 
In the speaker-independent stage, our networks are designed to capture generic relationships between texts and visual appearances. 
Unlike previous methods \cite{suwajanakorn2017synthesizing,taylor2017deep,fried2019text} that only synthesize and blend mouth region pixels, our method intends to generate holistic facial expression changes and head motions.
Hence, we design three networks to map input texts into animation parameters of the mouth, upper face and head pose respectively.
Furthermore, we employ a motion capture system to construct the correspondences between high-quality facial expressions as well as head motions and audios as our training data. Thus, our collected data can be used for training our speaker-independent networks effectively without requiring long-time talking videos of specified persons.

Since the animation parameters output by our speaker-independent networks are generic, we need to tailor the animation parameters to the specific input speaker to achieve convincing generated videos.
In the speaker-specific stage, we take the animation parameters as input and then exploit them to rig a given speaker's facial landmarks. 
In addition, we also develop an adaptive-attention network to adapt the rigged landmarks to the speaking characteristics of the specified person. 
In doing so, we only require a much shorter reference video (around 5 minutes) of the new speaker, instead of more than one hour speaker-specific videos often requested by previous methods \cite{suwajanakorn2017synthesizing,fried2019text}.

Overall, our method produces photo-realistic talking-head videos from a short reference video of a target performer. The generated videos also present rich details of the performer, such as realistic clothing, hair, and facial expressions.

\section{Related work}
\label{sec:relatedWork}

\subsection{Facial Animation Synthesis}

Facial animation synthesis pre-defines a 3D face model and generates the animation parameters to control the facial variation. 
LSTM \cite{hochreiter1997long} is widely used in facial animation synthesis for sequential modeling. 
Several works take BiLSTM \cite{pham2017speech}, CNN-LSTM \cite{pham2017end} or carefully-designed LSTM \cite{zhou2018visemenet} with regression loss, GAN loss \cite{sadoughi2019speech} or multi-task training strategy \cite{sadoughi2017joint} to synthesize full facial/mouth animation.
However, LSTM tends to work slower due to the sequential computation.
CNN is proven to have comparable ability to deal with sequential data \cite{bai2018empirical}. Some works employ CNN to animate mouth or full face from acoustic features \cite{karras2017audio,cudeiro2019capture} or time-aligned phonemes \cite{taylor2017deep}. 
Head animation synthesis focuses on synthesizing head pose from input speech. Some works direct regress head pose with BiLSTM \cite{Ding2015BLSTMNN,greenwood2018joint} or the encoder of transformer \cite{vaswani2017attention}. More precisely, head pose generation from speech is a one-to-many mapping, \citet{sadoughi2018novel} employ GAN \cite{goodfellow2014generative,mirza2014conditional,yu2019can,yu2019semantic} to retain the diversity.

\begin{figure*}
\centering
\includegraphics[width=0.90\textwidth]{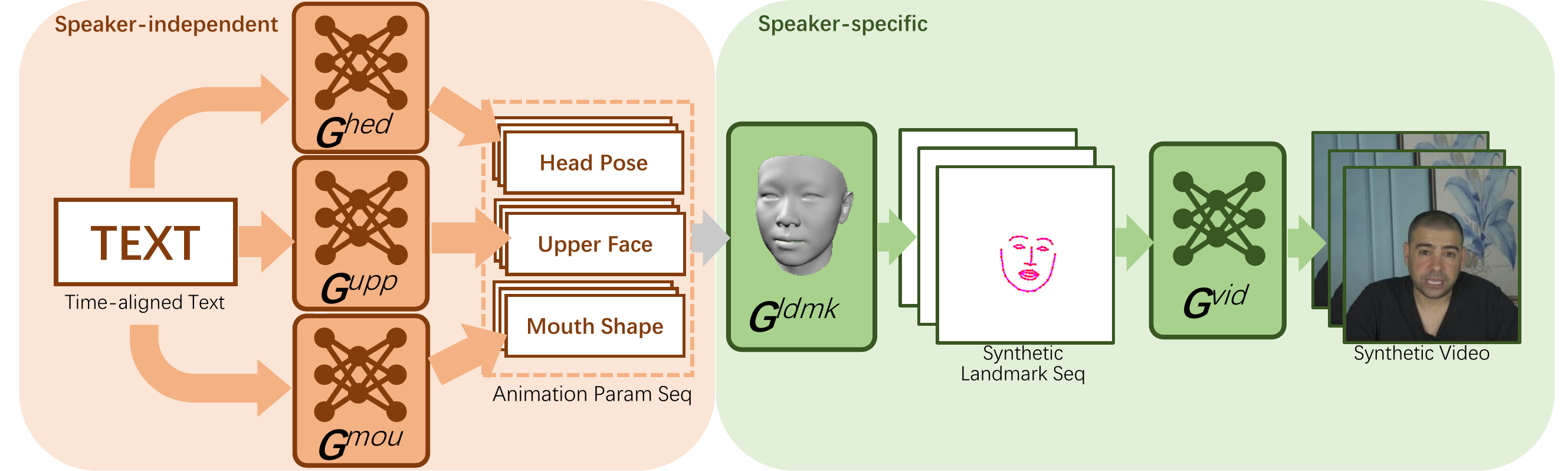}
\caption{Pipeline of our method. 
The speaker-independent stage takes the time-aligned text as input and generates head pose, upper face, and mouth shape animation parameters. The speaker-specific stage then produces synthetic talking-head videos from the animation parameters.}
\label{fig:pipeline}
\end{figure*}

\subsection{Face Video Synthesis}
\subsubsection{Audio-driven.}
Audio-driven face video synthesis directly generates 2D talking video from input audio.
Previous works \cite{vougioukas2019end,chen2018lip,zhou2019talking,wiles2018x2face,prajwal2020lip} utilize two sub-modules to compute face embedding feature and audio embedding feature for the target speaker, then fuse them as input to a talking-face generator. 
Another group of works decouple geometry generation and appearance generation into two stages. The geometry generation stage infers appropriate facial landmarks, which is taken as input by the appearance generation stage. 
Landmarks are inferred with speaker-specific model \cite{suwajanakorn2017synthesizing,dasspeech,zhou2020makelttalk} or linear principal components \cite{chen2019hierarchical,chen2020talking}.
\citet{thies2019neural} generate expression coefficients of a 3D Morphable Model (3DMM), then employ a neural renderer to generate photo-realistic images. 
\citet{fried2019text} infer expression parameters by searching and blending existing expressions of the reference video, then employ a recurrent neural network to generate the modified video. 
Although also taking text as input,
their method generates novel sentences inefficiently (10min-2h) due to the viseme search. 
Besides, both works fail to control the upper face and head pose to match the speech rhythm and emotion.

\subsubsection{Video-driven.}
Video-driven methods transfer expressions of one person to another.
Several works \cite{ha2019marionette,zeng2020realistic,song2019geometry,siarohin2019first} take a single image as the identity input. 
Other works take videos \cite{thies2015real,thies2018headon} as identity input to improve visual quality. 
\citet{thies2016face2face} reconstruct and renders a mesh model and fill in the inner mouth as output, the reconstructed face texture stays constant while talking. 
Some works directly generate 2D images with GAN instead of 3D rendering \cite{nirkin2019fsgan,zakharov2019few,wu2018reenactgan,thies2019deferred}.
\citet{kim2019neural} preserve the mouth motion style based on sequential learning on the unpaired data of the two speakers. 
Alternatively, our work generates paired mouth expression data to make the style learning easier. 
\citet{kim2018deep} also employs a 3DMM to render geometry information.
Instead of transferring existing expressions, our method generates new expressions from text. 
Furthermore, our method preserves the speaker's mouth motion style and designs an adaptive-attention network to obtain higher image resolution and better visual quality.

\section{Text-based Talking-head Generation}

Our framework takes the time-aligned text as input and outputs the photo-realistic talking-head video. It can be generalized to a specific speaker with about 5 minutes of his/her talking video (reference video). 
Figure \ref{fig:pipeline} illustrates the pipeline of our framework. 
Taking time-aligned text as input, $G^{mou}$, $G^{upp}$ and $G^{hed}$ separately generate speaker-independent animation parameters of mouth, upper face and head pose.
Instead of learning from the reference video, they take advantage of a Mocap dataset for higher accuracy.
Since a small error in geometry inference may lead to obvious artifacts in appearance inference, we introduce a 3D face module $G^{ldmk}$ to incorporate the head and facial expression parameters and convert them to speaker-specific facial landmark sequence.
Finally, $G^{vid}$ synthesizes the speaker-specific talking-head video according to the facial landmark sequence by rendering the texture of hair, face, upper torso and background.

\subsection{Mocap Dataset}
\label{sec:dataset}

\begin{figure}
 \includegraphics[width=0.45\textwidth]{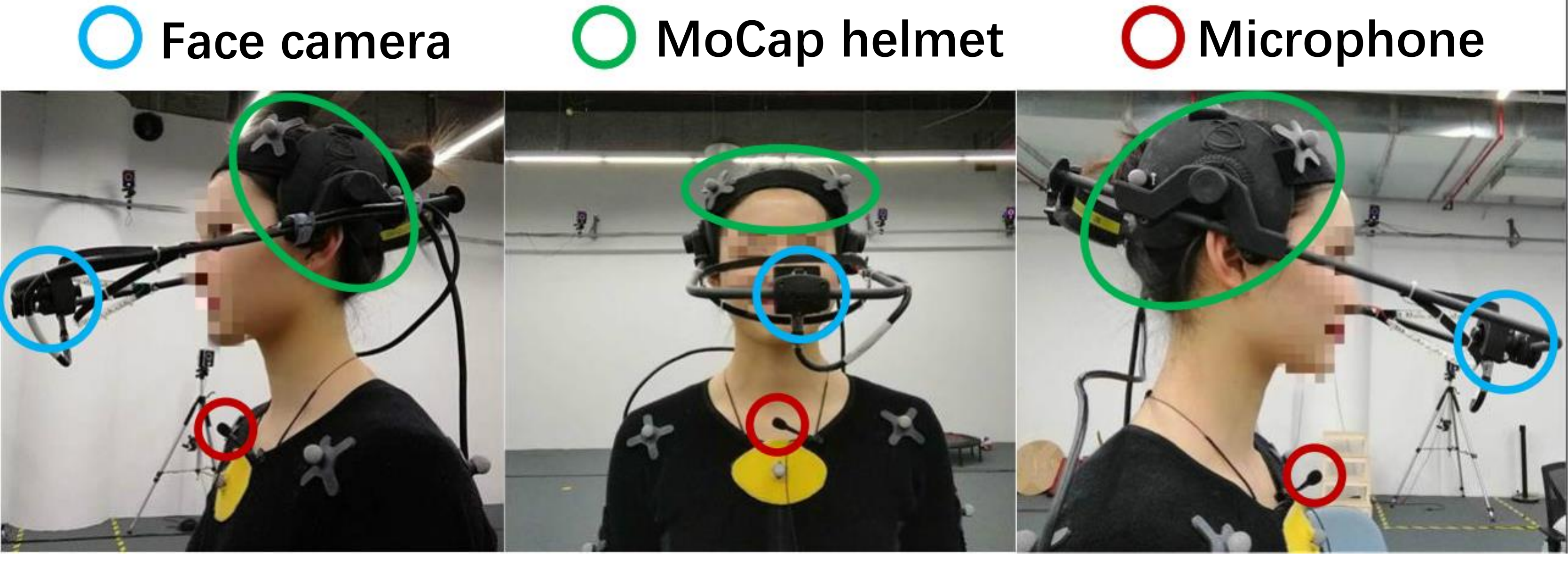}
 \caption{The collection of Mocap dataset. The recording is carried out by a professional actress wearing a helmet. Markers on the helmet offer information of head pose. The infrared camera attached to the helmet records accurate facial expressions.}
 \label{Fig: mocap}
\end{figure}

\begin{figure}
  \includegraphics[width=0.45\textwidth]{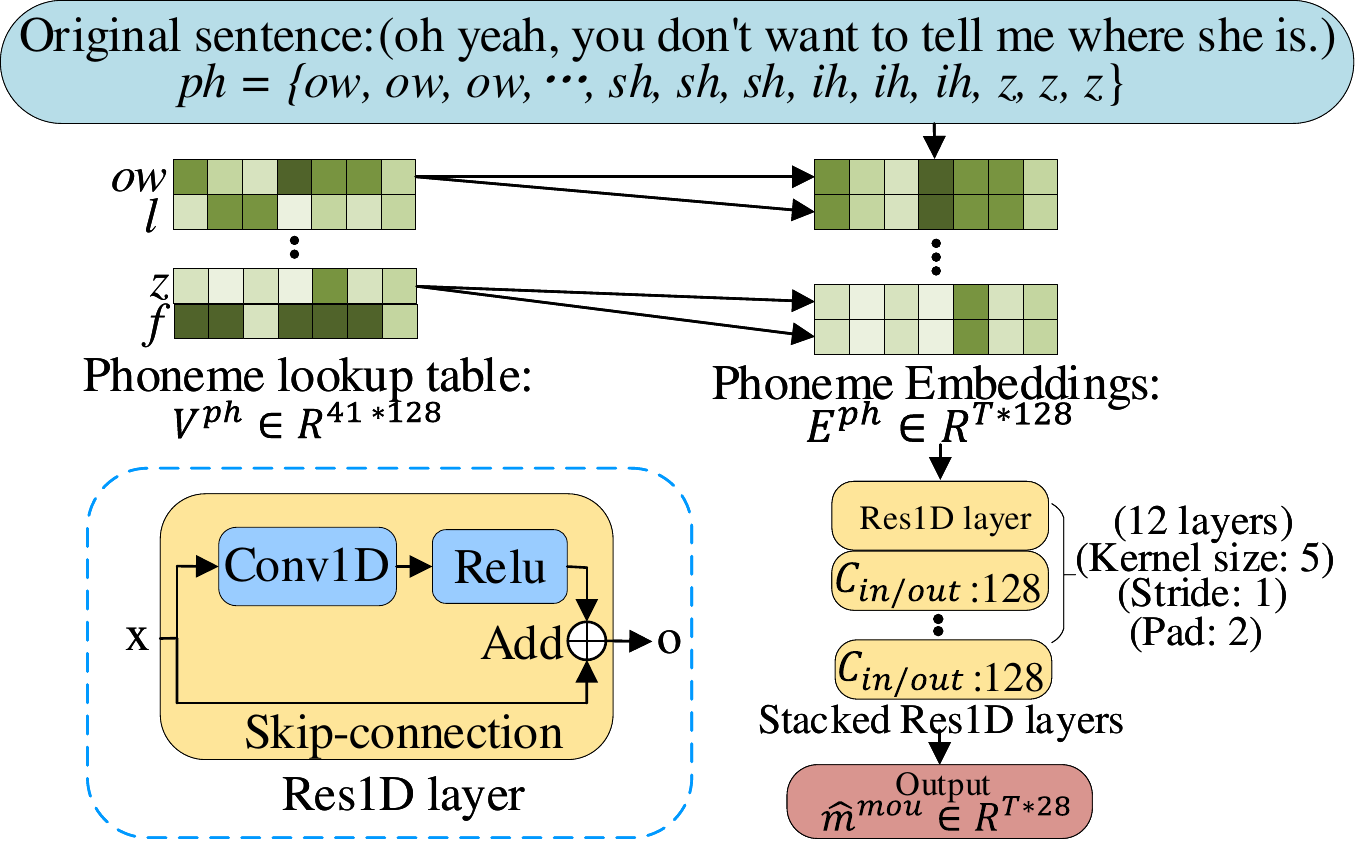} 
  \caption{Mouth animation generator.}
  \label{Fig:mouthModule}
\end{figure}

To obtain high-fidelity full facial expressions and head pose, we record an audiovisual dataset relying on a motion capture (Mocap) system\footnote{Dynamixyz, http://www.dynamixyz.com}\label{dy} shown in Figure \ref{Fig: mocap}. 
The collected data includes the mouth parameter sequence $m^{mou}=\{m^{mou}_t\}_{t=1}^{T}$ where $m^{mou}_t \in \mathbb{R}^{28}$, the upper face parameter sequence $m^{upp}=\{m^{upp}_t\}_{t=1}^{T}$ where $m^{upp}_t\in \mathbb{R}^{23}$ and the head pose parameter sequence $m^{hed}=\{m^{hed}_t\}_{t=1}^{T}$ where $m^{hed}_t\in \mathbb{R}^{6}$. 
$T$ is the length of frames in an utterance. 
$m^{mou}$ and $m^{upp}$ are defined as blendshape weights following the definition of Faceshift. Each blendshape stands for some part of the face movement, \eg eye-open, mouth-left.
We record 865 emotional utterances of a professional actress in English
($203$ surprise, $273$ anger, $255$ neutral and $134$ happiness), each of which lasts from 3 to 6 seconds. 
A time alignment analyzer
\footnote{qdreamer.com}
is employed to compute the duration of each phoneme and each word from audio. 
According to the alignment result, we represent the word sequence and phoneme sequence as $w=\{w_t\}_{t=1}^{T}$ and $ph=\{ph_t\}_{t=1}^{T}$ separately, where $w_t$ and $ph_t$ are the word and phoneme uttered at the $t$-th frame.
In this way, we build a high-fidelity Mocap dataset including $m^{mou}$, $m^{upp}$, $m^{hed}$, $w$ and $ph$, which is then used to train the speaker-independent generators.
Another Chinese dataset
(925 utterances from 3 to 6 seconds)
is similarly built.
Both datasets are released for research purposes\footnote{https://github.com/FuxiVirtualHuman/Write-a-Speaker}.

\subsection{Mouth Animation Generator}\label{sec:mouth} 
Since the mouth animation mainly contributes to uttering phonemes instead of semantic structures,
$G^{mou}$ learns a mapping from $ph$ to $m^{mou}$ ignoring $w$, as shown in Figure \ref{Fig:mouthModule}.  
The first step is to convert $ph$ from phoneme space into the embedding vectors $E^{ph}$ in a more flexible space.
We construct a trainable lookup table \cite{tang2014learning} $V^{ph}$ to meet the goal, which is randomly initialized and updated in the training stage.
Afterwards, The stacked Res1D layers take $E^{ph}$ as input and output synthetic mouth parameter sequence $\hat{m}^{mou}$ according to co-articulation effects.
We design the structure based on CNN instead of LSTM for the benefits of parallel computation. 

We apply $L_{1}$ loss and LSGAN loss \cite{LSGAN} for training $G^{mou}$. 
The $L_{1}$ loss is written as 
\begin{equation} 
L^{mou}_1 = \frac{1}{ T}\sum_{i=1}^{T}{(\|m_{i}^{mou} - \hat{m}_{i}^{mou}\|}_1),
\end{equation}
where $m_{i}^{mou}$ and $\hat{m}_{i}^{mou}$ are the real and generated vector of the $i$th frame separately. 
The adversarial loss is denoted as
\begin{equation} 
L^{mou}_{adv} = arg\min\limits_{G^{mou}}\max\limits_{D^{mou}} L_{GAN}(G^{mou},D^{mou}).
\end{equation}
Inspired by the idea of patch discriminator \cite{pix2pixgan}, $D^{mou}$ is applied on temporal trunks of blendshape which also consists of stacked Res1D layers.
The objective function is written as
\begin{equation}
\label{mouth-equation}
L(G^{mou}) =L^{mou}_{adv} + \lambda_{mou}L^{mou}_1.
\end{equation}

\begin{figure}
\includegraphics[width=0.45\textwidth]{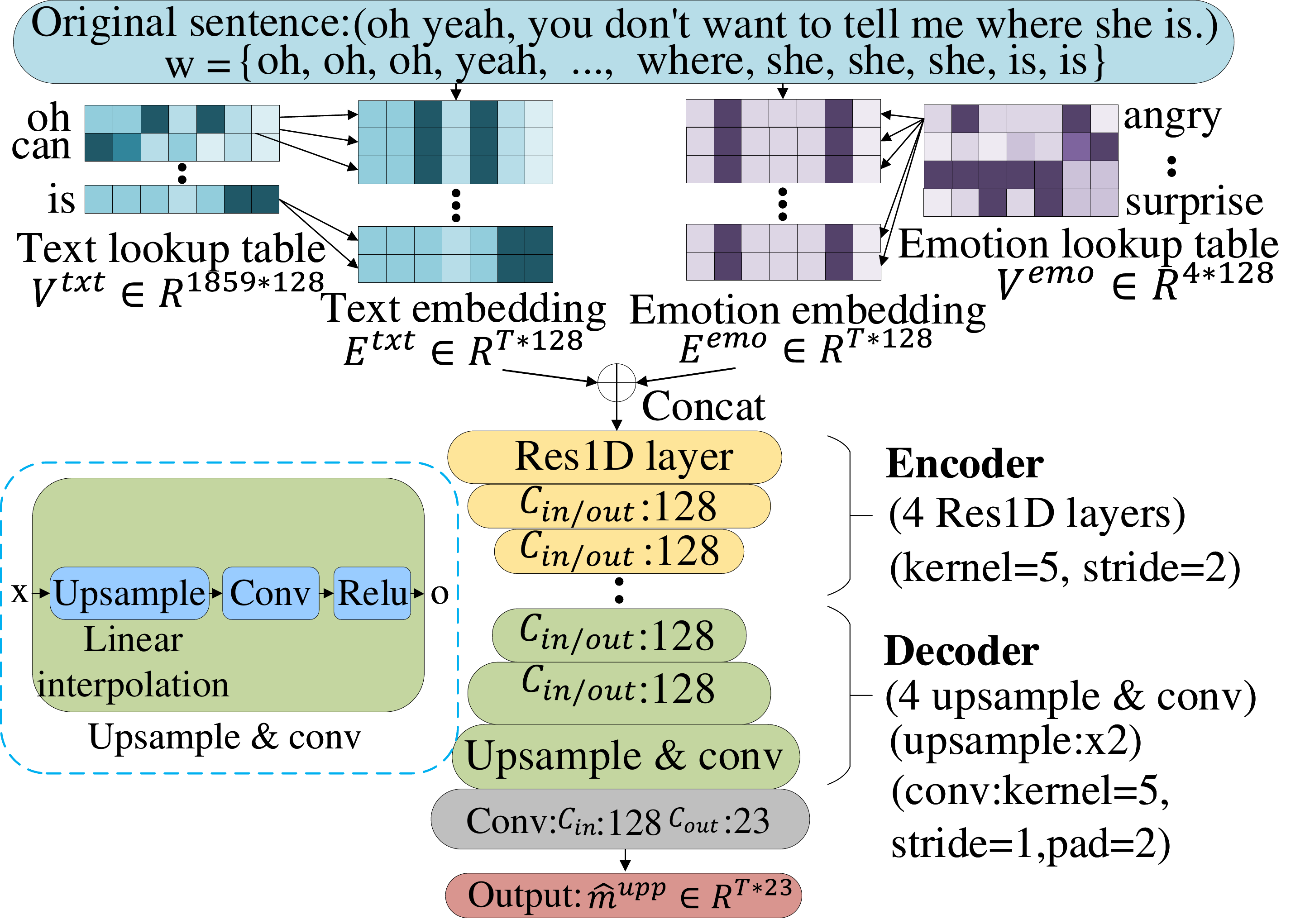} 
 \caption{Upper facial expression generator.} 
 \label{fig:encoder-decoder}
\end{figure}

\subsection{Upper Face/Head Pose Generators}
\label{sec:upper}

While mouth motions contribute to speech co-articulation, upper facial expressions and head motions tend to convey emotion, intention, and speech rhythm.
Therefore, $G^{upp}$ and $G^{hed}$ are designed to capture longer-time dependencies from $w$ instead of $ph$. They share the same network and differ from that of $G^{mou}$,
as illustrated in Figure \ref{fig:encoder-decoder}.
Similar to $V^{ph}$, a trainable lookup table $V^{txt}$ maps $w$ to embedding vectors $E^{txt}$.
In order to generate $m^{upp}$ with consistent emotion, 
an emotion label (surprise, anger, neutral, happiness) is either detected by a text sentiment classifier \cite{yang2019xlnet},
or explicitly assigned for the specific emotion type.
Another trainable lookup table $V^{emo}$ projects the emotion label to embedding vectors $E^{emo}$.
$E^{txt}$ and $E^{emo}$ are fed to an encoder-decoder network to synthesize ${m}^{upp}$.
Benefits from the large receptive field, the encoder-decoder structure captures long-time dependencies between words.

Since synthesizing $m^{upp}$ from text is a one-to-many mapping, the $L_{1}$ loss is replaced with SSIM loss \cite{wang2004image}. SSIM simulates the human visual perception and has benefit of extracting structural information. We extend SSIM to perform on each parameter respectively, namely SSIM-Seq loss, formulated as 
\begin{equation}\label{upper facial ssim}
\begin{split}
L^{upp}_{S} = 1 - \frac{1}{23}\sum_{i=1}^{23}\frac{(2 \mu_{i} \hat{\mu}_{i} + \delta_1) (2 cov_i+\delta_2))}{(\mu_{i}^{2} + \hat{\mu}_{i}^{2} + \delta_1)(\sigma_{i}^{2} + \hat{\sigma}_{i}^{2} + \delta_2))}
\end{split}.
\end{equation}
$\mu_i$/$\hat{\mu_i}$ and $\sigma_i$/$\hat{\sigma_i}$ represent the mean and standard deviation of the $i$ dimension of real/synthetic $m^{upp}$, and $cov_i$ is the covariance. 
$\delta_1$ and $\delta_2$ are two small constants.
The GAN loss is denoted as
\begin{equation} 
L^{upp}_{adv} = arg\min\limits_{G^{upp}}\max\limits_{D^{upp}} L_{GAN}(G^{upp},D^{upp}).
\end{equation}
where $D^{upp}$ shares the same structure with $D^{mou}$. 
The objective function is written as
\begin{equation}
\label{eyebrow-equation}
L(G^{upp}) =L^{upp}_{adv} + \lambda_{upp}L^{upp}_{S}.
\end{equation}

$G^{hed}$ shares the same network and loss but ignores $V^{emo}$ to generate $m^{hed}$, as the variation of head poses in different emotions is less significant than that of facial expressions.

\subsection{Style-Preserving Landmark Generator}
\label{sec:landmark}

$G^{ldmk}$ reconstructs the 3D face from the reference video, then drive it to obtain speaker-specific landmark images.  
A multi-linear 3DMM $U(s,e)$ is constructed with shape parameters $s\in \mathbb{R}^{60}$ and expression parameters $e\in \mathbb{R}^{51}$.
The linear shape basis are taken from LSFM \cite{booth2018lsfm} and scaled by the singular values. We sculpture 51 facial blendshapes on LSFM as the expression basis following the definition of Mocap dataset, so that \(e\) is consistent with \((m_t^{upp},m_t^{mou})\).
A 3DMM fitting method is employed to estimate \(s\) of the reference video.
Afterwards, we drive the speaker-specific 3D face with generated $\hat{m}^{hed}$, $\hat{m}^{mou}$ and $\hat{m}^{upp}$ to get the landmark image sequence. 
Our earlier experiments show that videos generated from the landmark images and rendered dense mesh are visually indifferent, we therefore choose landmark images to cut down a renderer.

Furthermore, speakers may use different mouth shapes to pronounce the same word, e.g. some people tend to open their mouths larger than others, and people are sensitive to the mismatched styles. 
Meanwhile, the generic $\hat{m}^{upp}$ and $\hat{m}^{hed}$ work fine among different people in practice.
Hence, we retarget $\hat{m}^{mou}$ to preserve the speaker's style while leaving $\hat{m}^{upp}$ and $\hat{m}^{hed}$ unchanged.
On one hand, we extract time-aligned text from the reference video and generate $\hat{m}^{mou}$ using $G^{mou}$. 
On the other hand, we estimate personalized $\breve{m}^{mou}$ from the reference video using 3DMM. 
In this way, we obtain paired mouth shapes pronouncing the same phonemes.
With the paired data, the style-preserving mapping from $\hat{m}^{mou}$ to $\breve{m}^{mou}$ is easily learnt. A two-layer fully-connected network with MSE loss works well in our experiments. We use the mapped $\breve{m}^{mou}$ to produce the landmark images.

\subsection{Photo-realistic Video Generator}

$G^{vid}$ produces the talking-head video $\{\hat{I}_t\}_{t=1}^{T}$ frame by frame from the landmark images. $\hat{I}_t$ depicts the speaker's full facial expression, hair, head and upper torso poses, and the background at the $t$-th frame.
Considering the high temporal coherence, we construct the conditional space-time volume $V$ as input of $G^{vid}$ by stacking the landmark images in a temporal sliding window of length 15. 

\label{sec:photoRealGenerator}
\begin{figure}
  \includegraphics[width=0.45\textwidth]{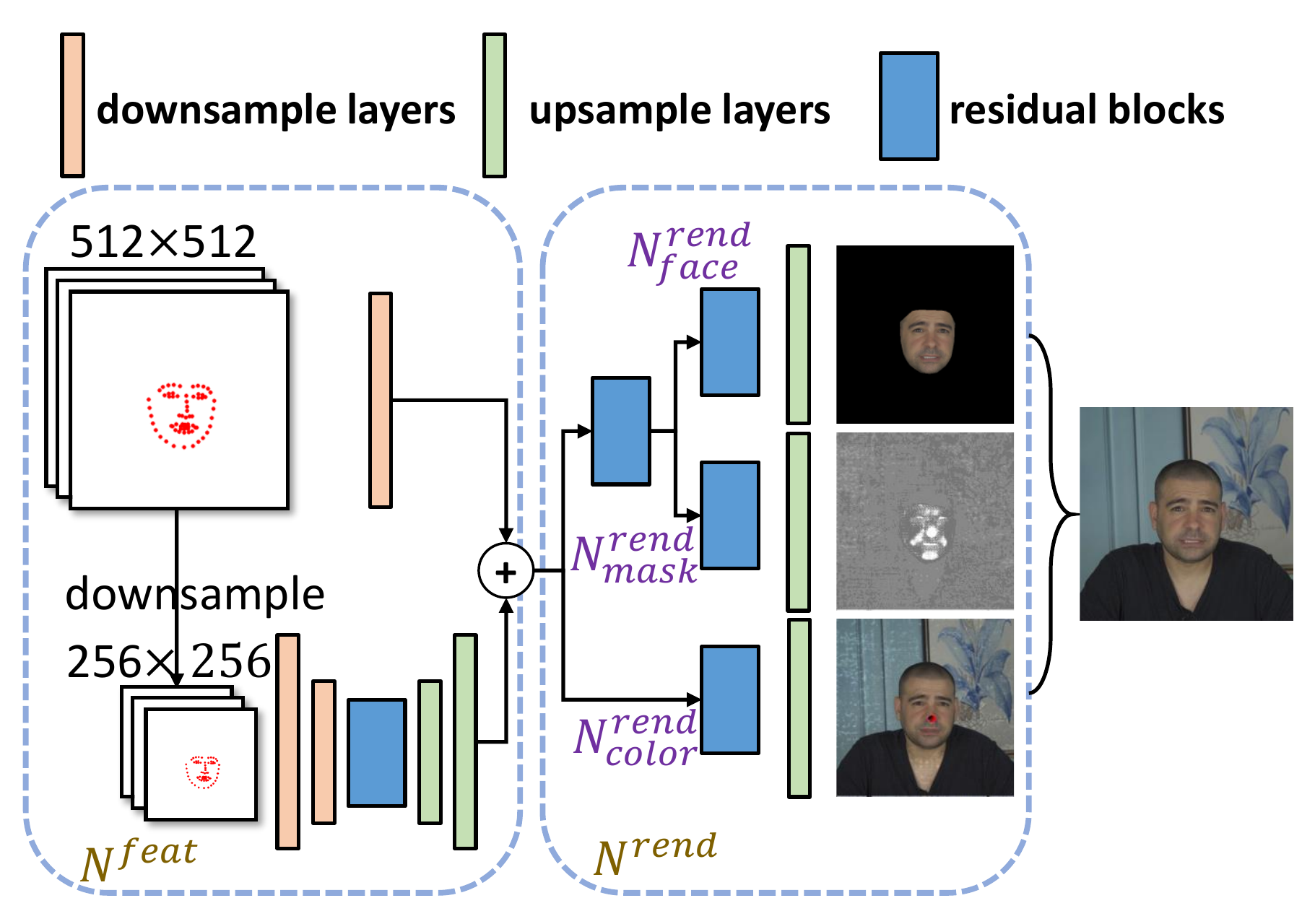}
  \vspace{-1em}
  \caption{Photo-realistic video generator.}
  \vspace{-1em}
  \label{fig:landmark2video}
\end{figure}

Although typical image synthesis networks \cite{pix2pixgan,pix2pixhd,yu2016ultra,yu2017face,yu2017hallucinating} are able to produce reasonable head images, their outputs tend to be blurry on areas with high-frequency movements, especially the eye and mouth regions.
The possible explanation is that the movements of eye and mouth are highly correlated with landmarks while the torso pose and background are less, so it is not the best solution to treat all parts as a whole. 
Motivated by the observation, we design an adaptive-attention structure.
As shown in Figure \ref{fig:landmark2video}, $G^{vid}$ is composed by a feature extraction network ${N^{feat}}$ and self-attention rendering network ${N^{rend}}$.
To extract features from high resolution landmark images, $N^{feat}$ consists of two pathways of different input scales. The extracted features of the two pathways are element-wise summed.
${N^{rend}}$ renders talking-head images from the latent features. To model the different correlations of body parts, we design a composite of three parallel sub-networks.
${N^{rend}_{face}}$ produces the target face $\hat{I}^{face}$. ${N^{rend}_{clr}}$ is expected to compute the global color map $\hat{I}^{color}$, with hair, upper body, background and so on. ${N^{rend}_{mask}}$ produces the adaptive-attention fusion mask $M$ that focus on the high-frequency-motion regions. 
The final generated image $\hat{I}_t$ is given by
\begin{equation} \label{generated_image}
\hat{I}_t = M*\hat{I}^{face}+(1-M)*\hat{I}^{color}.
\end{equation}

\begin{figure}
 \includegraphics[width=0.45\textwidth]{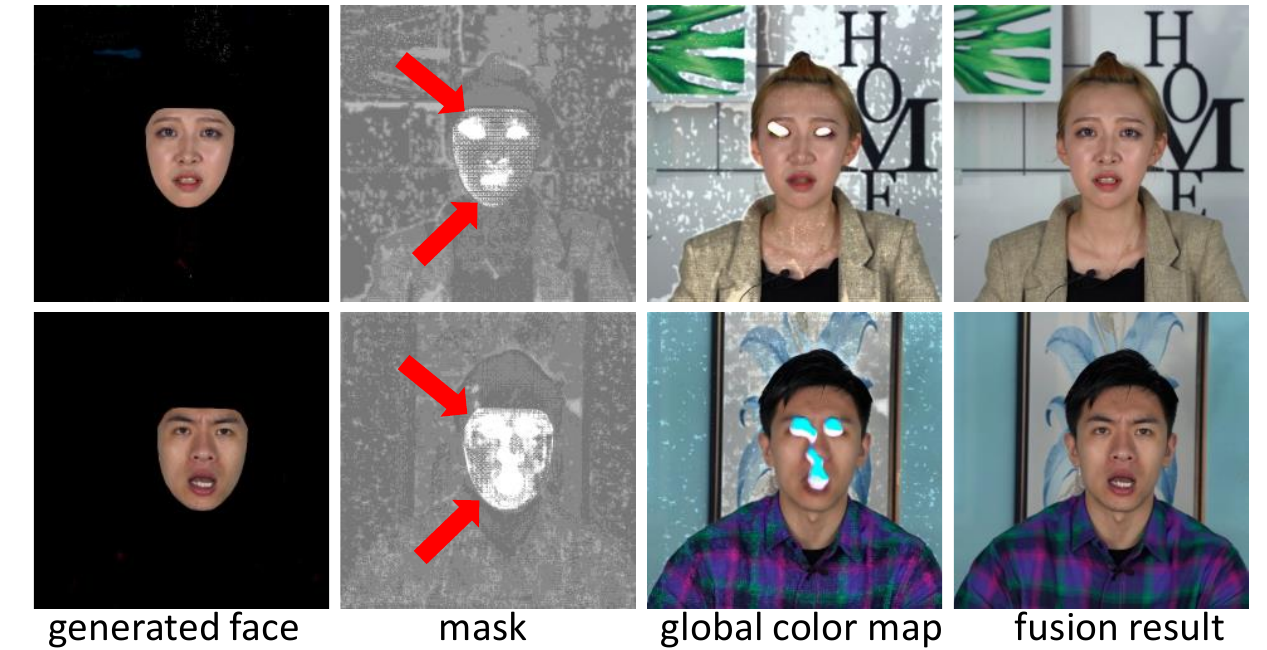} 
 \caption{Sample outputs of our photo-realistic video generator. It shows that the adaptive-attention mask is able to distinguish the region of mouth and eyes from other regions.} 
 \label{fig:attention}
\end{figure}

Figure \ref{fig:attention} shows the details of our attention mask. 

We follow the discriminators of pix2pixHD \cite{pix2pixhd}, consisting of 3 multi-scale discriminators $D^{vid}_1$, $D^{vid}_2$ and $D^{vid}_3$. 
The inputs of them are $\hat{I}_t$/$I_t$ and $V$, where $I_t$ is the real frame.
The adversarial loss is defined as:
\begin{equation} \label{adversarial_loss}
L^{vid}_{adv}=\min\limits_{G^{vid}}\max\limits_{D^{vid}_1,D^{vid}_2,D^{vid}_3}\sum_{i=1}^3L_{GAN}(G^{vid},D^{vid}_i),
\end{equation}

To capture the fine facial details we adopt the perceptual loss \cite{percepture_loss}, following~\citet{yu2018face}
\begin{equation} \label{perceptual_loss}
L_{perc} = \sum_{i=1}^{n}\frac{1}{W_iH_iC_i}{\|F_i(I_t)-F_i(\hat{I}_t)\|}_1,
\end{equation}
where $F_i \in \mathbb{R}^{W_i \times H_i \times C_i}$ is the feature map of the $i$-th layer of VGG-19 \cite{VGG}. 
Matching both lower-layer and higher-layer features guides the generation network to learn both fine-grained details and a global part arrangement. Besides, we use $L_1$ loss to supervise the generated $\hat{I}_{face}$ and $\hat{I}_t$:
\begin{equation} \label{img_loss}
L^{img}_1=\|I_t-\hat{I}_t\|_1,\\
L^{face}_1=\|I^{face}_t-\hat{I}^{face}_t\|_1.
\end{equation}
$I_{face}$ is cropped from $I_t$ according to the detected landmarks \cite{baltrusaitis2018openface}.

The overall loss is defined as:
\begin{equation} \label{G_loss}
L(G^{vid})=\alpha L_{perc}+\beta L^{img}_1+\gamma L^{face}_1+L^{vid}_{adv}.
\end{equation}

\section{Experiments and Results}

We implement the system using PyTorch on a single GTX 2080Ti. 
The training of the speaker-independent stage takes 3 hours on the Mocap dataset. The training of the speaker-specific stage takes one day on a 5 mins' reference video.
Our method produces videos of $512\times 512$ resolution at 5 frames per second.
More implementation details are introduced in the supplementary material.
We compare the proposed method with state-of-the-art audio/video driven methods, and evaluate the effectiveness of the submodules. Video comparisons are shown in the supplementary video.

\subsection{Comparison to Audio-driven Methods}

\begin{figure}
\includegraphics[width=0.47\textwidth]{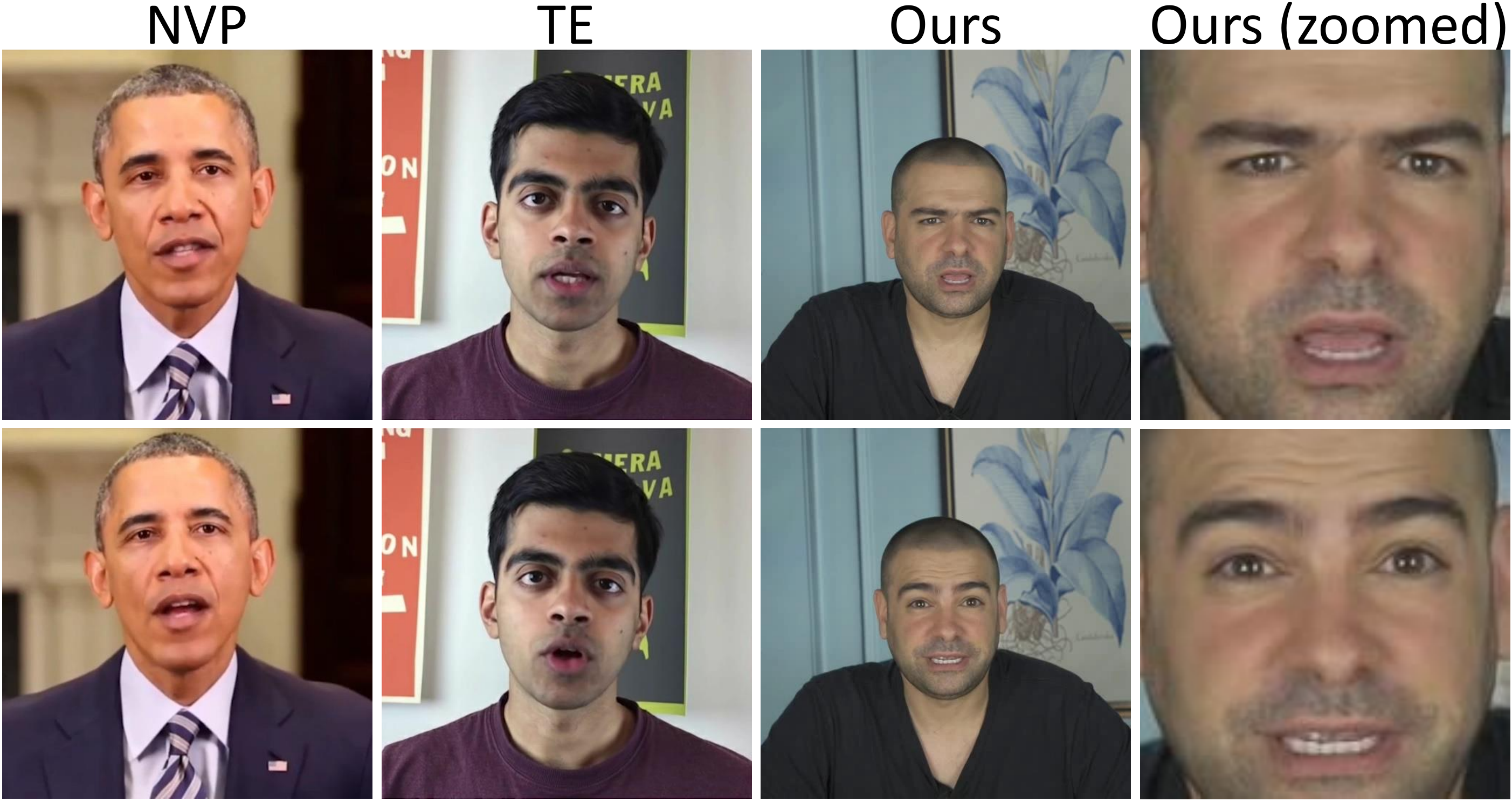} 
 \caption{Comparison with NVP and TE. Our approach matches the sentiment and rhythm of emotional audios.} 
 \label{fig:Comparison_TE_NVP}
\end{figure}

We first compare our method with Neural Voice Puppetry (NVP) \cite{thies2019neural} and Text-based Editing (TE) \cite{fried2019text}, which achieve state-of-the-art visual quality by replacing and blending mouth region pixels of the reference video. 
As shown in Figure \ref{fig:Comparison_TE_NVP},
while achieving similar visual quality on non-emotional speech, our method additionally controls the upper face and head motion to match the sentiment and rhythm of emotional audios. 
In contrast, NVP and TE do not have mechanisms to model sentiments of audio.

\begin{figure}
\includegraphics[width=0.47\textwidth]{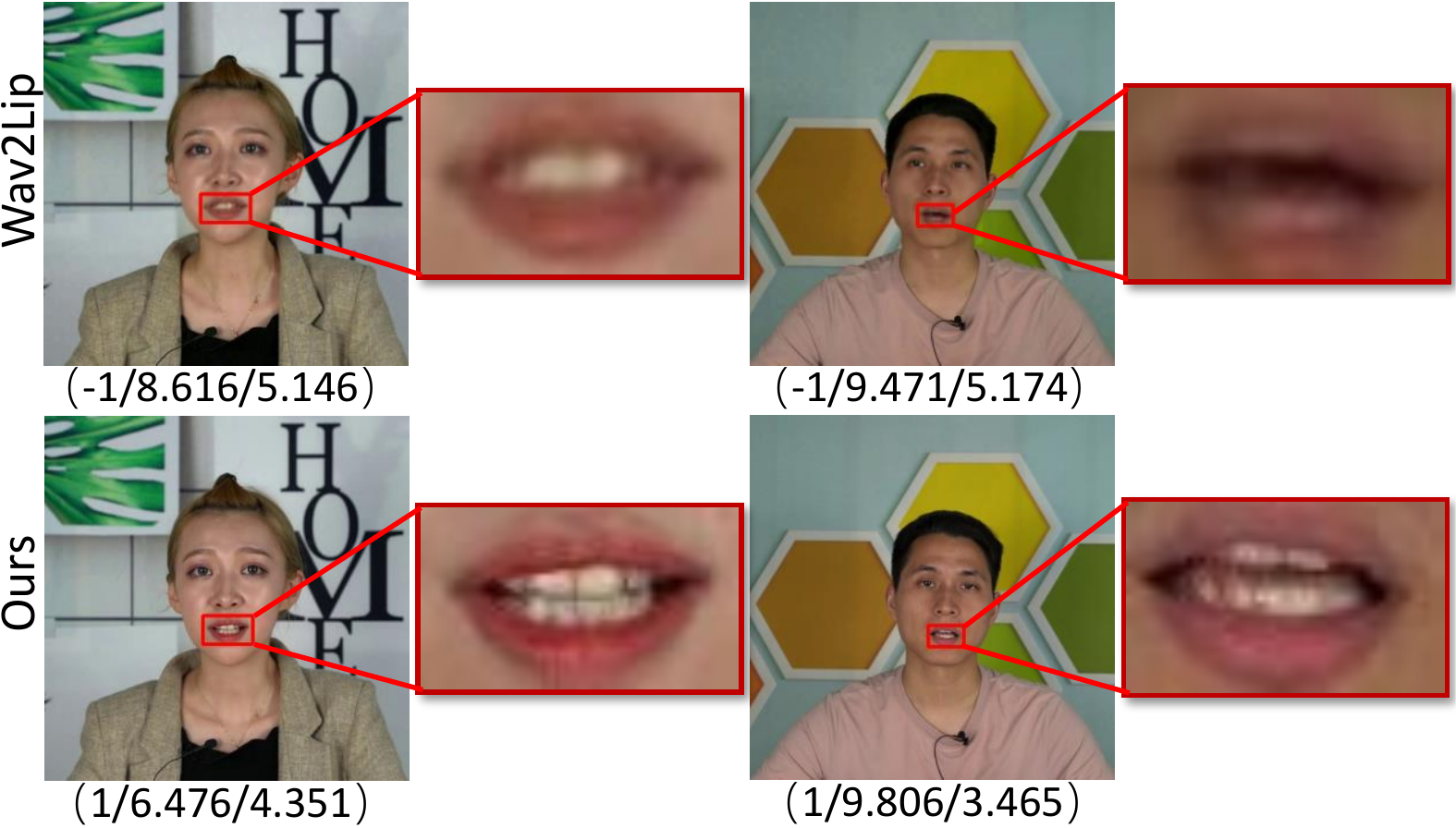} 
 \caption{Comparison with Wav2Lip. Metrics of SyncNet are listed below (offset/distance/confidence). }
 \label{fig:compare_with_wav2lip}
\end{figure}

\begin{figure}
\includegraphics[width=0.47\textwidth]{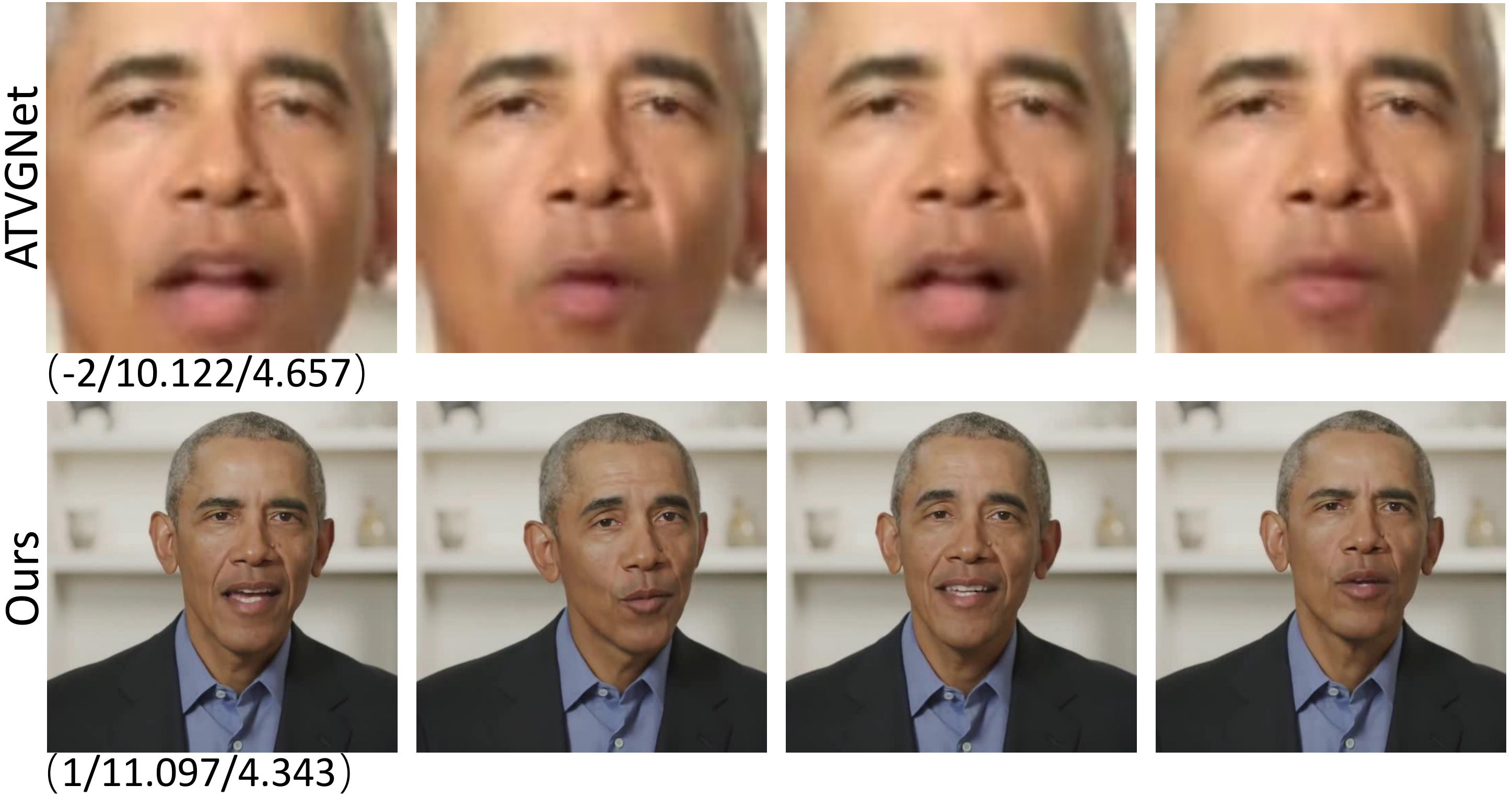} 
 \caption{Comparison with ATVGNet. Metrics of SyncNet are listed below (offset/distance/confidence).} 
 \label{fig:Comparison_Few_Shot_Audio}
\end{figure}

We then compare our method with Wav2Lip \cite{prajwal2020lip} in Figure \ref{fig:compare_with_wav2lip}, which only requires a reference video of a few seconds. Metrics of SyncNet \cite{Chung16a} are listed below each image. Although their method produces accurate lip shapes from audio, we can observe the obvious artifacts in the inner mouth.
Our method is compared to ATVGNet \cite{chen2019hierarchical} in Figure \ref{fig:Comparison_Few_Shot_Audio}, which produces talking head videos from a single image. Their method focuses on low resolution cropped front faces while our method generates high-quality full head videos. Considering their method learns identity information from one image instead of a video, the visual quality gap is as expected.

\subsection{Comparison to Video-driven Methods}

\begin{figure}
\centering
\includegraphics[width=0.47\textwidth]{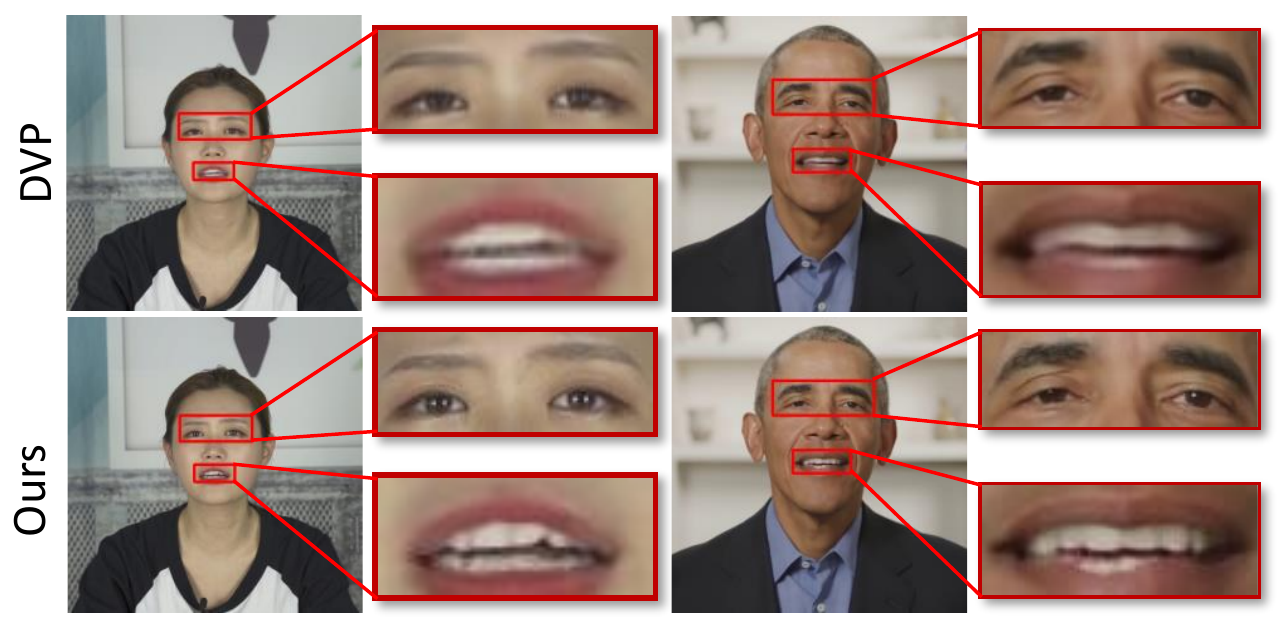}
\caption{Comparison with DVP.}
\label{fig:com_with_dvp}
\end{figure}

We also compare our method with Deep Video Portrait (DVP) \cite{kim2018deep}, whose original intention is expression transfer. We reproduce DVP and replace their detected animation parameters with our generated animation parameters for fair comparison. Results are shown in Figure \ref{fig:com_with_dvp}. Although our method uses sparse landmarks instead of rendered dense mesh, we synthesize better details on mouth and eye regions.

\subsection{Evaluation of Submodules}

In order to evaluate $G^{mou}$ and $G^{upp}$, we reproduce the state-of-the-art facial animation synthesis works \cite{karras2017audio,pham2017end,sadoughi2017joint,taylor2017deep,cudeiro2019capture,sadoughi2019speech}. 
For fair comparison, their input features and network structures are retained and the output is replaced with facial expression parameters.
To further evaluate the loss terms, we additionally conduct an experiment by removing the GAN loss in Equation \ref{mouth-equation} and \ref{eyebrow-equation} (Ours w/o GAN).
The groundtruth test data is selected from the high accuracy Mocap dataset.
For mouth parameters, we measure MSE of $m^{mou}$ and lips landmark distance (LMD) on 3D face mesh. LMD is measured on 3D face mesh instead of 2D images to avoid the effect of head pose variation. 
For upper face parameters, we measure SSIM of $m^{upp}$.
Results are shown in Table \ref{comparasion-text2blendshape}. 
Both $G^{mou}$ and $G^{upp}$ perform better than the above methods.

\begin{table}
\small
\setlength{\tabcolsep}{1.7mm}{
\begin{tabular}{l|ccc}
\hline
  & MSE$\downarrow$ & LMD$\downarrow$ & SSIM$\uparrow$  \\ 
\hline 
  \cite{karras2017audio} &88.75    &0.0690        &0.0931                  \\ 
\hline
  \cite{pham2017end} &109.02    &0.0742        &0.0889                  \\ 
\hline
  \cite{sadoughi2017joint} &103.34    &0.0721      &0.0793                  \\
\hline
\cite{taylor2017deep} &89.59     &0.0699       & $-$                  \\ 
\hline
\cite{cudeiro2019capture} &91.22    &0.0713      &$-$                  \\ 
\hline
\cite{sadoughi2019speech} &89.29    &0.0694        &$-$                  \\ \hline
Ours w/o GAN & 89.21     &0.0693       & 0.1879                \\ \hline
Ours & \textbf{87.24}     &\textbf{0.0684}       & \textbf{0.2655}                 \\ \hline
\end{tabular}
}
\caption{Quantitative evaluattion $G^{mou}$ and $G^{upp}$.}
\label{comparasion-text2blendshape}
\end{table}

\begin{figure}
 \includegraphics[width=0.48\textwidth]{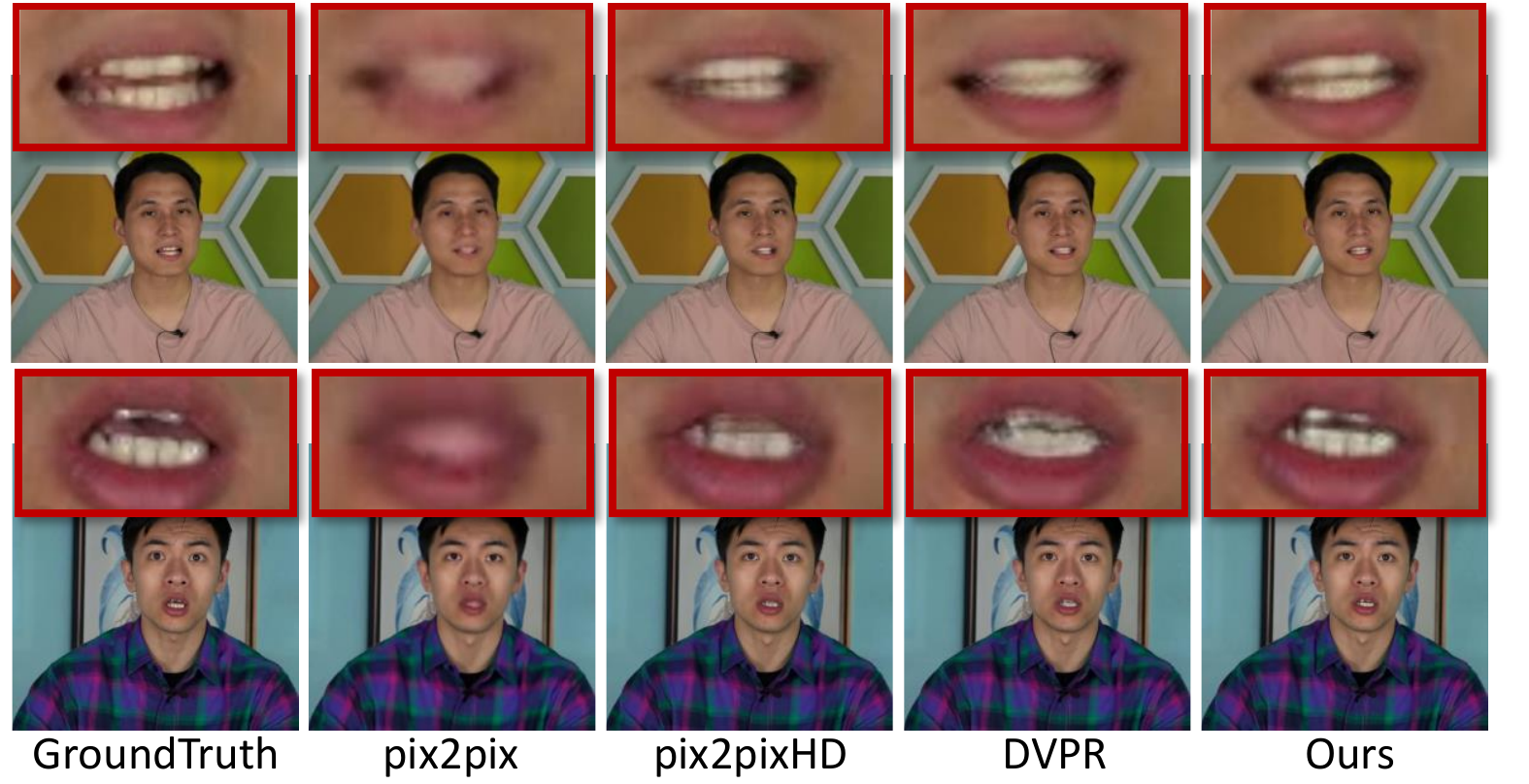}
 \vspace{-2em}
 \caption{Comparison of $G^{vid}$ and the state-of-the-arts.}
 \label{Fig: validation}
 \vspace{-1em}
\end{figure}

To prove the superiority of $G^{vid}$, 
we compare $G^{vid}$ with pix2pix \cite{pix2pixgan}, pix2pixHD \cite{pix2pixhd} and photo-realistic rendering network of DVP (denoted as DVPR). 
To evaluate the results, we apply multiple metrics including SSIM, Fréchet Inception Distance (FID) \cite{FID}, Video Multimethod Assessment Fusion (VMAF) and Cumulative Probability of Blur Detection (CPBD) \cite{narvekar2011no}.
For fair comparison, we take the same space-time volume as the input of all networks and train them on the same datasets. 
Table \ref{tab:landmark2img-metric} shows the quantitative results, and
Figure \ref{Fig: validation} shows the qualitative comparison. 
Our approach is able to produce higher quality of images, especially on teeth and eyes regions.

\begin{table}
    \centering
    \small
    \setlength{\tabcolsep}{1.8mm}{
    \begin{tabular}{c|c|cccc}
    \hline
    \multicolumn{2}{c|}{} & SSIM$\uparrow$ & FID$\downarrow$ & VMAF$\uparrow$ & CPBD$\uparrow$ \\
    \hline
    \multirow{4}*{ID1} & pix2pix & 0.9466  & 0.1279 & 62.75 & 0.1233 \\
    ~ & pix2pixHD & 0.9455 & 0.02711 & 65.42 & 0.2517 \\
    ~ & DVPR & 0.9371 & 0.02508 & 57.75 & 0.2607 \\
    ~ & Ours & \textbf{0.9490} &\textbf{0.01452} & \textbf{66.68} & \textbf{0.2682} \\
    \hline
    \multirow{4}*{ID2} & pix2pix & 0.9026 & 0.04360 & 60.32 & 0.1083 \\
    ~ & pix2pixHD & 0.8998 & 0.01883 & 60.37 & 0.2572 \\
    ~ & DVPR & 0.9031 & 0.009456 & 62.27 & 0.2859 \\
    ~ & Ours & \textbf{0.9042} & \textbf{0.003252} & \textbf{63.76} & \textbf{0.2860} \\
    \hline
    \multirow{4}*{ID3}  & pix2pix & 0.9509 & 0.04631 & 72.15 & 0.2467 \\
    ~ & pix2pixHD & 0.9499 & 0.005940 & 74.64 & 0.3615 \\
    ~ & DVPR & 0.9513 & 0.005232 & 71.12 & 0.3642 \\
    ~ & Ours & \textbf{0.9514} & \textbf{0.003262} & \textbf{74.76} & \textbf{0.3661} \\
    \hline
    \end{tabular}}
    \caption{Quantitative evaluation of $G^{vid}$.}
    \label{tab:landmark2img-metric}
\end{table}

\subsection{Ablation Study}

\begin{figure}
\centering
\includegraphics[width=0.49\textwidth]{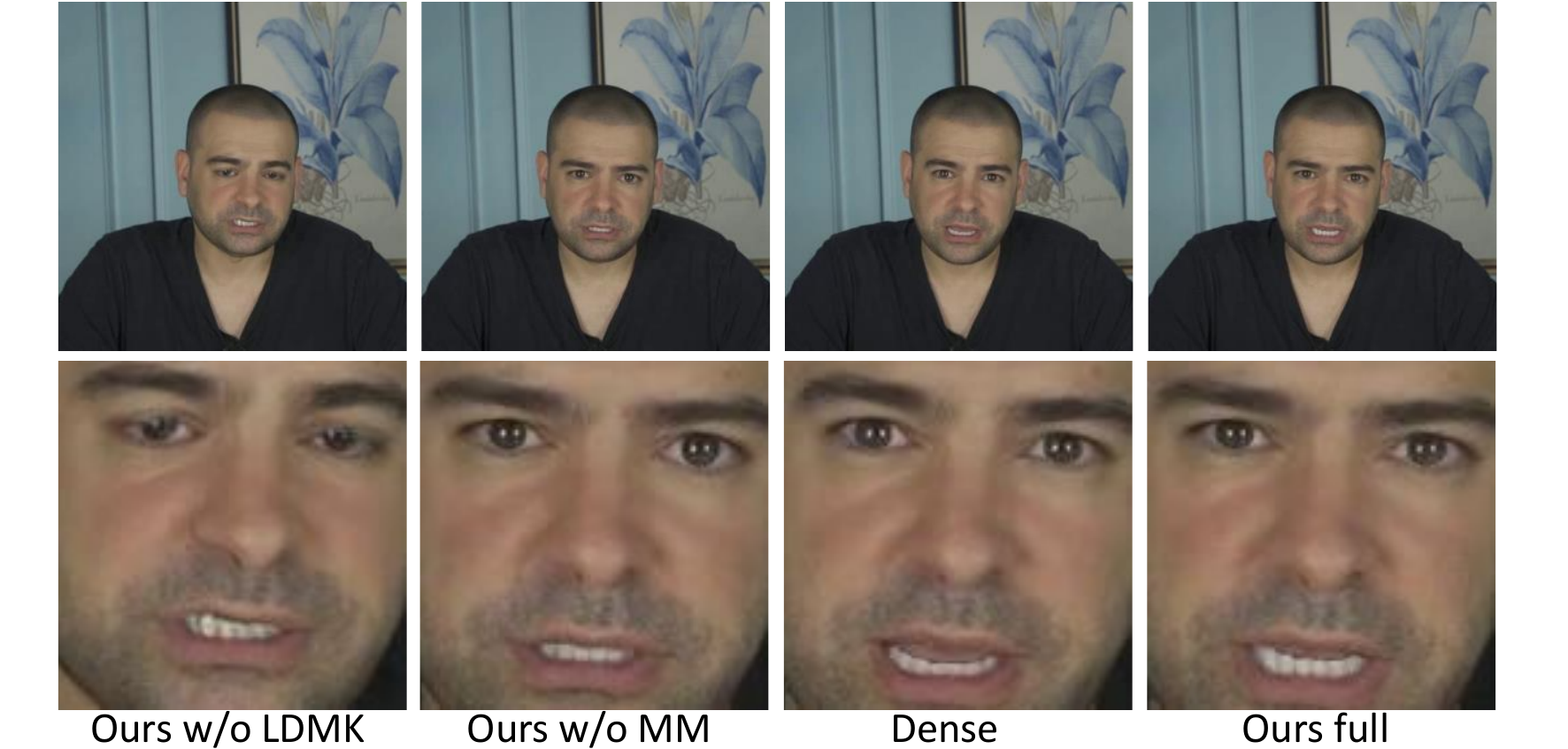}
\caption{Results of different conditions.}
\label{fig:ablation}
\end{figure}

\begin{figure}
\centering
\includegraphics[width=0.45\textwidth]{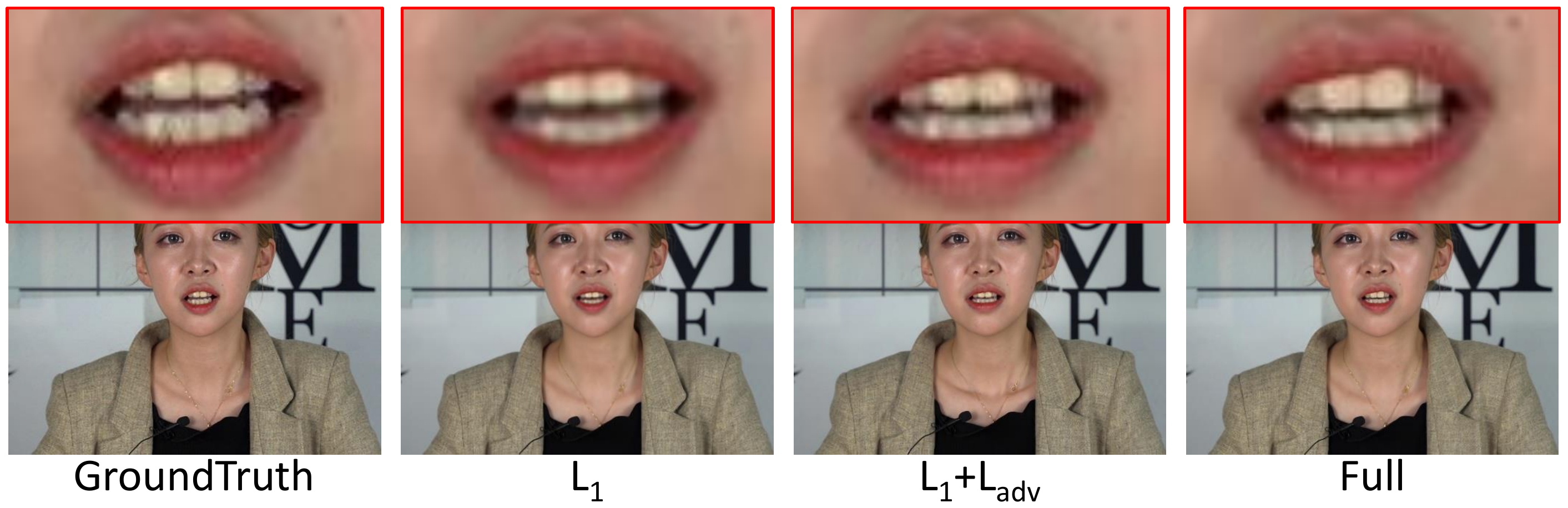}
\caption{Results from different loss terms of $G^{vid}$.}
\label{fig:ablation_Gvid}
\end{figure}

We perform an ablation study to evaluate other components of our framework, results are shown in Figure \ref{fig:ablation}. We remove or replace several submodules to construct the input of $G^{vid}$.
The first condition removes $G^{ldmk}$ and directly input animation parameters to $G^{vid}$ (w/o LDMK). Due to the lack of explicit geometry constraint, the output contains some twisted and jittered face regions.
The second condition uses $G^{ldmk}$ but removes the mouth style mapping (w/o MM). The speaker in the output video opens his mouth smaller than in the reference video for pronunciation, preserving the mismatched style of the actress of the Mocap dataset.
The third condition additionally replaces the sparse landmarks with dense 3D face mesh (dense). The visual quality of the output is visually indifferent with that of our method, indicating that the sparse geometry constraint is good enough for $G^{vid}$. 
Figure \ref{fig:ablation_Gvid} shows another ablation study to evaluate the effectiveness of each loss terms in $G^{vid}$. All loss terms contribute to the visual quality.

\subsection{User Study}
We further conduct an online user study to evaluate the quality of the output videos. We compare our method with groundtruth videos (GT),
ours with extracted $m^{upp}$ and $m^{hed}$ from reference videos instead of generated (Ours w/o E\&H),
DVP, Wav2Lip. We generate 5 sentences of the same speaker in the same resolution for each method, to obtain $5 \times 5 = 25$ video clips. 
The audios are extracted from the reference video.
60 participants are asked to rate the realism of each video clip. Results are listed in Table \ref{tab:user_study} ($60\times5=300$ ratings for each method). 
Only $91\%$ of GT are judged as real, indicating that participants are overcritical when trying to detect synthesized videos.
Even with the comparison of real videos, our results are judged as real in $52\%$ of the cases. our method outperforms all compared methods significantly ($p<0.001$) in both mean score and 'judged as real' proportion. 
Results of 'Ours w/o E\&H' contain expression and head motion that do not match the speech sentiment and rhythm.
The difference between 'Ours' and 'Ours w/o E\&H' validates the effectiveness of our generated emotional upper face expressions and rhythmic head motions. 
The main reason of lower scores of DVP and Wav2Lip may be the artifacts in the inner mouth.

\begin{table}
\centering
{
\setlength{\tabcolsep}{0.9mm}{
\small
\begin{tabular}{c|c c c c c c c}
\hline
         & 1   &   2   & 3   & 4   & 5   &  Mean & 'real'(4+5) \\ \hline
GT       & 0\%   &1\%   &8\%   &37\%  &54\%  &4.45  &   91.3\%  \\ 
Wav2Lip  & 11\%  &29\%  &28\%  &30\%  &3\%   &2.85  &  32.3\%  \\ 
DVP      & 11\%  &29\%  &42\%  &17\%  &1\%   &2.67  & 18.0\%  \\ 
Ours w/o E\&H & 3\%   &22\%  &43\%  &26\%  &7\%   &3.13  & 33.0\%  \\ 
Ours     & 1\%   &9\%  &38\%  &37\%  &15\%  & \textbf{3.56} &
\textbf{51.7\%}  \\ \hline
\end{tabular}}}
\caption{Results of the user study. Participants are asked to rate the videos by 1-completely fake, 2-fake, 3-uncertain, 4-real, 5-completely real. Percentage numbers are rounded.}
\label{tab:user_study}
\end{table}

\begin{figure}
\centering
\includegraphics[width=0.45\textwidth]{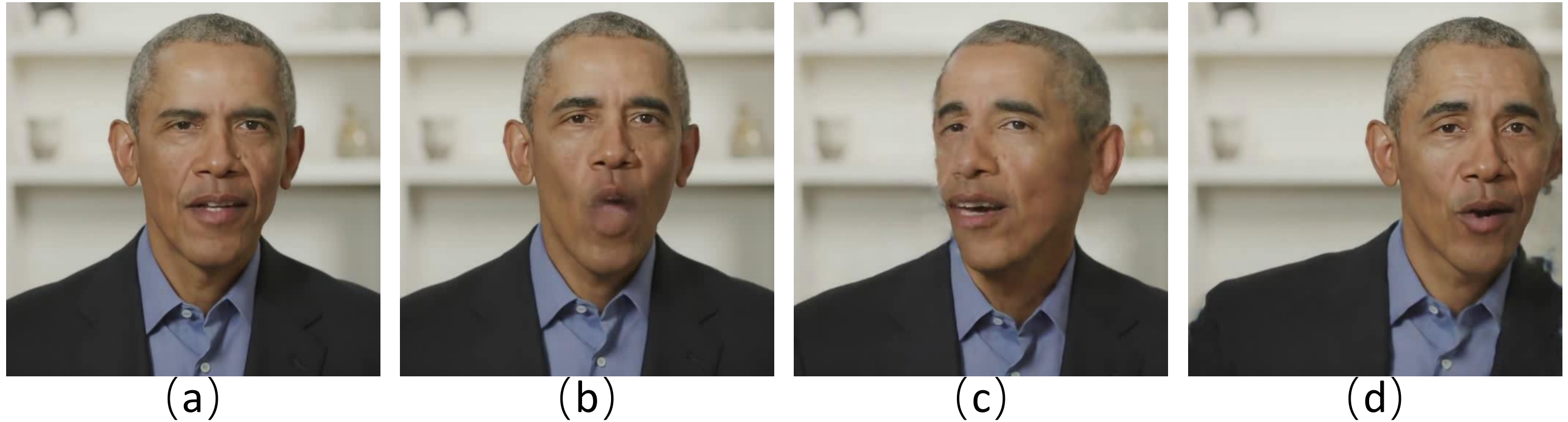}
\caption{Failure cases from extreme parameters, including (a) upper facial expression; (b) mouth expression; (c) head rotation; (d) head translation.}
\label{fig:failure_case}
\end{figure}

\section{Limitations}

Our work has several limitations. 
The proposed method takes advantage of a high-quality Mocap dataset. 
Our approach is restricted to produce speakers uttering in English or Chinese, because we have only captured Mocap datasets of the two languages. 
The amount of Mocap data is also insufficient to capture more detailed correspondences of motions and semantic and syntactic structures of text input. 
In the near future, we will record Mocap data of more languages and release them for the research purpose. 
Our rendering network cannot tackle with dynamic background and complex upper torso movements, such as shrugging, swinging arms, hunching back, extreme head poses an so on. 
The generated videos will degenerate if the expected expression or head motion is beyond the scope of the reference video.
The effect of emotion is ignored on the generated lip and head animations. 
Figure \ref{fig:failure_case} shows some failure cases.
In the future, we will be devoted to addressing the above problems.

\section{Conclusion}
This paper presents a text-based talking-head video generation framework.
The synthesized video displays the emotional full facial expressions, rhythmic head motions, the upper torso movements, and the background. 
The generation framework can be adapted to a new speaker with 5 minutes of his/her reference video. 
Our method is evaluated through a series of experiments, including qualitative evaluation and quantitative evaluation. 
The evaluation results show that our method can generate high-quality photo-realistic talking-head videos and outperforms the state-of-the-art. 
To the best of our knowledge, our work is the first to produce full talking-head videos with emotional facial expressions and rhythmic head movements from the time-aligned text representation.

\section{Ethical Consideration}
To ensure proper use, we firmly require that any result created using our algorithm must be marked as synthetic with watermarks. 
As part of our responsibility, for the positive applications, we intend to share our dataset and source code so that it can not only encourage efforts in detecting manipulated video content but also prevent the abuse. 
Our text-based talking head generation work can contribute to many positive applications, and we encourage further discussions and researches regarding the fair use of synthetic content.

\section{Appendix}

\subsection{3DMM Fitting}

We select \(N_k=30\) keyframes and aim to find the optimal variable set 
\(\boldsymbol{X}=(s, m^{hed}_1, e_{1}, ... , m^{hed}_{N_k}, e_{N_k})\),
where \(m^{hed}_k\) and \(e_{k}\) are the pose and expression parameters of the \(k\)-th keyframe. 
We focus on the \(N_l=68\) facial landmark consistency by minimizing the following energy function:
\begin{equation}
\begin{split}
F(X) = \sum_{k=1}^{N_k}( \sum_{i=1}^{N_l}Dis(p_{k,i}, P(U(s,e_{k})^{(i)},m^{hed}_{k})) \\+ \lambda_e \left \| e_{k} \right \|_{2}^{2}) + \lambda_s \left \| s \right \|_{2}^{2},
\end{split}
\end{equation}
where \(p_{k,i}\) is the coordinate of the \(i\)-th landmark detected from the \(k\)-th keyframe \cite{baltrusaitis2018openface}, and \(U^{(i)}\) is the \(i\)-th 3D landmark on mesh \(U\).
\(P(U^{(i)},m^{hed}_{k})\) projects \(U^{(i)}\) with pose \(m^{hed}_{k}\) into image coordinates. 
\(Dis(\cdot, \cdot)\) measures the distance of the projected mesh landmark and the detected image landmark. 
The regularization weights are set to \(\lambda_e=10^{-4}\) and \(\lambda_s=10^{-4}\). 
We employ the Levenberg-Marquard algorithm for the optimization.

\subsection{Network Structure and Training}

The size of $V^{ph}$ is $41 \times 128$, where $41$ is the number of phonemes and $128$ is the phoneme embedding size. The row vectors of $E^{ph} \in \mathbb{R}^{T \times 128}$ are picked up from $V^{ph}$ according to the phoneme indexes.
The size of $V^{txt}$ is 
$1859 \times 128$, where $1859$ means $1858$
words and one 'unknown' flag for all other words, and $128$ is the word embedding size. The size of $V^{emo}$ is $4 \times 128$. Each row of $V^{emo}$ represents an emotion embedding.
$N^{rend}_{face}$ and $N^{rend}_{mask}$ share the first 3 residual blocks. 
The top layer of ${N^{rend}_{face}}$/${N^{rend}_{clr}}$ is activated by tanh and that of ${N^{rend}_{mask}}$ is done by sigmoid.
The loss weights are set to $\lambda_{mou}=50$, $\lambda_{upp}=100$, $\alpha=10$, $\beta=100$, and $\gamma=100$.
We use the Adam \cite{adam} optimizer for all networks. 
For training $G^{mou}$, $G^{upp}$ and $G^{hed}$, we set $\beta_1=0.5,\beta_2=0.99,\epsilon=10^{-8}$, batch size of 32, and set the initial learning rate as 0.0005 for the generators and 0.00001 for the discriminators. The learning rates of $G^{mou}$ stay fixed in the first 400 epoches and linearly decay to zero within another 400 epoches. 
The learning rates of $G^{upp}$ and $G^{hed}$ keep unchanged in the first 50 epoches and linearly decay to zero within another 50 epoches. 
We randomly select $1\sim3$ words as 'unknown' in each sentence to improve the performance from limited training data.
For training $G^{vid}$, we set $\beta_1=0.5,\beta_2 = 0.999, \epsilon=10^{-8}$, batch size of 3, and initial learning rate of 0.0002 with linear decay to 0.0001 within 50 epochs.

\bibliography{aaai.bib}

\begin{thebibliography}{65}
\providecommand{\natexlab}[1]{#1}
\providecommand{\url}[1]{\texttt{#1}}
\providecommand{\urlprefix}{URL }
\expandafter\ifx\csname urlstyle\endcsname\relax
  \providecommand{\doi}[1]{doi:\discretionary{}{}{}#1}\else
  \providecommand{\doi}{doi:\discretionary{}{}{}\begingroup
  \urlstyle{rm}\Url}\fi

\bibitem[{Bai, Kolter, and Koltun(2018)}]{bai2018empirical}
Bai, S.; Kolter, J.~Z.; and Koltun, V. 2018.
\newblock An empirical evaluation of generic convolutional and recurrent
  networks for sequence modeling.
\newblock \emph{arXiv preprint arXiv:1803.01271} .

\bibitem[{Baltrusaitis et~al.(2018)Baltrusaitis, Zadeh, Lim, and
  Morency}]{baltrusaitis2018openface}
Baltrusaitis, T.; Zadeh, A.; Lim, Y.~C.; and Morency, L.-P. 2018.
\newblock Openface 2.0: Facial behavior analysis toolkit.
\newblock In \emph{FG 2018}, 59--66. IEEE.

\bibitem[{Booth et~al.(2018)Booth, Roussos, Ponniah, Dunaway, and
  Zafeiriou}]{booth2018lsfm}
Booth, J.; Roussos, A.; Ponniah, A.; Dunaway, D.; and Zafeiriou, S. 2018.
\newblock Large scale 3D morphable models.
\newblock \emph{IJCV} 126(2-4): 233--254.

\bibitem[{Chen et~al.(2020)Chen, Cui, Liu, Li, Kou, Xu, and
  Xu}]{chen2020talking}
Chen, L.; Cui, G.; Liu, C.; Li, Z.; Kou, Z.; Xu, Y.; and Xu, C. 2020.
\newblock Talking-head Generation with Rhythmic Head Motion.
\newblock \emph{arXiv preprint arXiv:2007.08547} .

\bibitem[{Chen et~al.(2018)Chen, Li, K~Maddox, Duan, and Xu}]{chen2018lip}
Chen, L.; Li, Z.; K~Maddox, R.; Duan, Z.; and Xu, C. 2018.
\newblock Lip movements generation at a glance.
\newblock In \emph{ECCV}, 520--535.

\bibitem[{Chen et~al.(2019)Chen, Maddox, Duan, and Xu}]{chen2019hierarchical}
Chen, L.; Maddox, R.~K.; Duan, Z.; and Xu, C. 2019.
\newblock Hierarchical cross-modal talking face generation with dynamic
  pixel-wise loss.
\newblock In \emph{CVPR}, 7832--7841.

\bibitem[{Chou et~al.(2018)Chou, Yeh, Lee, and Lee}]{taida}
Chou, J.-c.; Yeh, C.-c.; Lee, H.-y.; and Lee, L.-s. 2018.
\newblock Multi-target Voice Conversion without Parallel Data by Adversarially
  Learning Disentangled Audio Representations.
\newblock \emph{Interspeech} 501--505.

\bibitem[{Chung and Zisserman(2016)}]{Chung16a}
Chung, J.~S.; and Zisserman, A. 2016.
\newblock Out of time: automated lip sync in the wild.
\newblock In \emph{ACCVW}.

\bibitem[{Cudeiro et~al.(2019)Cudeiro, Bolkart, Laidlaw, Ranjan, and
  Black}]{cudeiro2019capture}
Cudeiro, D.; Bolkart, T.; Laidlaw, C.; Ranjan, A.; and Black, M.~J. 2019.
\newblock Capture, learning, and synthesis of 3D speaking styles.
\newblock In \emph{CVPR}, 10101--10111.

\bibitem[{Das et~al.(2020)Das, Biswas, Sinha, and Bhowmick}]{dasspeech}
Das, D.; Biswas, S.; Sinha, S.; and Bhowmick, B. 2020.
\newblock Speech-driven Facial Animation using Cascaded GANs for Learning of
  Motion and Texture.
\newblock In \emph{ECCV}.

\bibitem[{Ding, Zhu, and Xie(2015)}]{Ding2015BLSTMNN}
Ding, C.; Zhu, P.; and Xie, L. 2015.
\newblock BLSTM neural networks for speech driven head motion synthesis.
\newblock In \emph{INTERSPEECH}, 3345--3349.

\bibitem[{Ekman(1997)}]{ekman1997face}
Ekman, R. 1997.
\newblock \emph{What the face reveals: Basic and applied studies of spontaneous
  expression using the Facial Action Coding System (FACS)}.
\newblock Oxford University Press, USA.

\bibitem[{Fried et~al.(2019)Fried, Tewari, Zollh{\"o}fer, Finkelstein,
  Shechtman, Goldman, Genova, Jin, Theobalt, and Agrawala}]{fried2019text}
Fried, O.; Tewari, A.; Zollh{\"o}fer, M.; Finkelstein, A.; Shechtman, E.;
  Goldman, D.~B.; Genova, K.; Jin, Z.; Theobalt, C.; and Agrawala, M. 2019.
\newblock Text-based editing of talking-head video.
\newblock \emph{TOG} 38(4): 1--14.

\bibitem[{Goodfellow et~al.(2014)Goodfellow, Pouget-Abadie, Mirza, Xu,
  Warde-Farley, Ozair, Courville, and Bengio}]{goodfellow2014generative}
Goodfellow, I.; Pouget-Abadie, J.; Mirza, M.; Xu, B.; Warde-Farley, D.; Ozair,
  S.; Courville, A.; and Bengio, Y. 2014.
\newblock Generative adversarial nets.
\newblock In \emph{NeurIPS}, 2672--2680.

\bibitem[{Greenwood, Matthews, and Laycock(2018)}]{greenwood2018joint}
Greenwood, D.; Matthews, I.; and Laycock, S.~D. 2018.
\newblock Joint Learning of Facial Expression and Head Pose from Speech.
\newblock In \emph{Interspeech}, 2484--2488.

\bibitem[{Ha et~al.(2020)Ha, Kersner, Kim, Seo, and Kim}]{ha2019marionette}
Ha, S.; Kersner, M.; Kim, B.; Seo, S.; and Kim, D. 2020.
\newblock MarioNETte: Few-shot Face Reenactment Preserving Identity of Unseen
  Targets.
\newblock In \emph{AAAI}, volume~34, 10893--10900.

\bibitem[{Heusel et~al.(2017)Heusel, Ramsauer, Unterthiner, Nessler, and
  Hochreiter}]{FID}
Heusel, M.; Ramsauer, H.; Unterthiner, T.; Nessler, B.; and Hochreiter, S.
  2017.
\newblock Gans trained by a two time-scale update rule converge to a local nash
  equilibrium.
\newblock In \emph{NeurIPS}, 6626--6637.

\bibitem[{Hochreiter and Schmidhuber(1997)}]{hochreiter1997long}
Hochreiter, S.; and Schmidhuber, J. 1997.
\newblock Long short-term memory.
\newblock \emph{Neural computation} 9(8): 1735--1780.

\bibitem[{Isola et~al.(2017)Isola, Zhu, Zhou, and Efros}]{pix2pixgan}
Isola, P.; Zhu, J.; Zhou, T.; and Efros, A.~A. 2017.
\newblock Image-to-image translation with conditional adversarial networks.
\newblock In \emph{CVPR}, 1125--1134.

\bibitem[{Johnson, Alahi, and Fei-Fei(2016)}]{percepture_loss}
Johnson, J.; Alahi, A.; and Fei-Fei, L. 2016.
\newblock Perceptual losses for real-time style transfer and super-resolution.
\newblock In \emph{ECCV}, 694--711. Springer.

\bibitem[{Karras et~al.(2017)Karras, Aila, Laine, Herva, and
  Lehtinen}]{karras2017audio}
Karras, T.; Aila, T.; Laine, S.; Herva, A.; and Lehtinen, J. 2017.
\newblock Audio-driven facial animation by joint end-to-end learning of pose
  and emotion.
\newblock \emph{TOG} 36(4): 94.

\bibitem[{Kim et~al.(2019)Kim, Elgharib, Zollh{\"o}fer, Seidel, Beeler,
  Richardt, and Theobalt}]{kim2019neural}
Kim, H.; Elgharib, M.; Zollh{\"o}fer, M.; Seidel, H.-P.; Beeler, T.; Richardt,
  C.; and Theobalt, C. 2019.
\newblock Neural style-preserving visual dubbing.
\newblock \emph{TOG} 38(6): 1--13.

\bibitem[{Kim et~al.(2018)Kim, Garrido, Tewari, Xu, Thies, Nie{\ss}ner,
  P{\'e}rez, Richardt, Zollh{\"o}fer, and Theobalt}]{kim2018deep}
Kim, H.; Garrido, P.; Tewari, A.; Xu, W.; Thies, J.; Nie{\ss}ner, M.;
  P{\'e}rez, P.; Richardt, C.; Zollh{\"o}fer, M.; and Theobalt, C. 2018.
\newblock Deep video portraits.
\newblock \emph{TOG} 37(4): 1--14.

\bibitem[{Kingma and Ba(2014)}]{adam}
Kingma, D.~P.; and Ba, J. 2014.
\newblock Adam: A method for stochastic optimization.
\newblock \emph{arXiv preprint arXiv:1412.6980} .

\bibitem[{Mao et~al.(2017)Mao, Li, Xie, Lau, Wang, and Paul~Smolley}]{LSGAN}
Mao, X.; Li, Q.; Xie, H.; Lau, R.~Y.; Wang, Z.; and Paul~Smolley, S. 2017.
\newblock Least squares generative adversarial networks.
\newblock In \emph{ICCV}, 2794--2802.

\bibitem[{Mignault and Chaudhuri(2003)}]{Mignault2003}
Mignault, A.; and Chaudhuri, A. 2003.
\newblock The Many Faces of a Neutral Face: Head Tilt and Perception of
  Dominance and Emotion.
\newblock \emph{J. Nonverbal Behav.} 27: 111--132.

\bibitem[{Mirza and Osindero(2014)}]{mirza2014conditional}
Mirza, M.; and Osindero, S. 2014.
\newblock Conditional generative adversarial nets.
\newblock \emph{arXiv preprint arXiv:1411.1784} .

\bibitem[{Narvekar and Karam(2011)}]{narvekar2011no}
Narvekar, N.~D.; and Karam, L.~J. 2011.
\newblock A no-reference image blur metric based on the cumulative probability
  of blur detection (CPBD).
\newblock \emph{TIP} 20(9): 2678--2683.

\bibitem[{Nirkin, Keller, and Hassner(2019)}]{nirkin2019fsgan}
Nirkin, Y.; Keller, Y.; and Hassner, T. 2019.
\newblock Fsgan: Subject agnostic face swapping and reenactment.
\newblock In \emph{ICCV}, 7184--7193.

\bibitem[{Pham, Cheung, and Pavlovic(2017)}]{pham2017speech}
Pham, H.~X.; Cheung, S.; and Pavlovic, V. 2017.
\newblock Speech-driven 3D facial animation with implicit emotional awareness:
  A deep learning approach.
\newblock In \emph{CVPRW}, 80--88.

\bibitem[{Pham, Wang, and Pavlovic(2017)}]{pham2017end}
Pham, H.~X.; Wang, Y.; and Pavlovic, V. 2017.
\newblock End-to-end learning for 3d facial animation from raw waveforms of
  speech.
\newblock \emph{arXiv preprint arXiv:1710.00920} .

\bibitem[{Prajwal et~al.(2020)Prajwal, Mukhopadhyay, Namboodiri, and
  Jawahar}]{prajwal2020lip}
Prajwal, K.; Mukhopadhyay, R.; Namboodiri, V.~P.; and Jawahar, C. 2020.
\newblock A Lip Sync Expert Is All You Need for Speech to Lip Generation In The
  Wild.
\newblock In \emph{MM}, 484--492.

\bibitem[{Sadoughi and Busso(2016)}]{sadoughi2016head}
Sadoughi, N.; and Busso, C. 2016.
\newblock Head Motion Generation with Synthetic Speech: A Data Driven Approach.
\newblock In \emph{INTERSPEECH}, 52--56.

\bibitem[{Sadoughi and Busso(2017)}]{sadoughi2017joint}
Sadoughi, N.; and Busso, C. 2017.
\newblock Joint learning of speech-driven facial motion with bidirectional
  long-short term memory.
\newblock In \emph{IVA}, 389--402. Springer.

\bibitem[{Sadoughi and Busso(2018)}]{sadoughi2018novel}
Sadoughi, N.; and Busso, C. 2018.
\newblock Novel realizations of speech-driven head movements with generative
  adversarial networks.
\newblock In \emph{ICASSP}, 6169--6173. IEEE.

\bibitem[{Sadoughi and Busso(2019)}]{sadoughi2019speech}
Sadoughi, N.; and Busso, C. 2019.
\newblock Speech-driven expressive talking lips with conditional sequential
  generative adversarial networks.
\newblock \emph{IEEE Transactions on Affective Computing} .

\bibitem[{Siarohin et~al.(2019)Siarohin, Lathuili{\`e}re, Tulyakov, Ricci, and
  Sebe}]{siarohin2019first}
Siarohin, A.; Lathuili{\`e}re, S.; Tulyakov, S.; Ricci, E.; and Sebe, N. 2019.
\newblock First order motion model for image animation.
\newblock In \emph{NeurIPS}, 7137--7147.

\bibitem[{Simonyan and Zisserman(2014)}]{VGG}
Simonyan, K.; and Zisserman, A. 2014.
\newblock Very deep convolutional networks for large-scale image recognition.
\newblock \emph{arXiv preprint arXiv:1409.1556} .

\bibitem[{Song et~al.(2019)Song, Cao, Song, Hu, and He}]{song2019geometry}
Song, L.; Cao, J.; Song, L.; Hu, Y.; and He, R. 2019.
\newblock Geometry-aware face completion and editing.
\newblock In \emph{AAAI}, volume~33, 2506--2513.

\bibitem[{Suwajanakorn, Seitz, and
  Kemelmacher-Shlizerman(2017)}]{suwajanakorn2017synthesizing}
Suwajanakorn, S.; Seitz, S.~M.; and Kemelmacher-Shlizerman, I. 2017.
\newblock Synthesizing obama: learning lip sync from audio.
\newblock \emph{TOG} 36(4): 1--13.

\bibitem[{Tang et~al.(2014)Tang, Wei, Yang, Zhou, Liu, and
  Qin}]{tang2014learning}
Tang, D.; Wei, F.; Yang, N.; Zhou, M.; Liu, T.; and Qin, B. 2014.
\newblock Learning sentiment-specific word embedding for twitter sentiment
  classification.
\newblock In \emph{ACL}, 1555--1565.

\bibitem[{Taylor et~al.(2017)Taylor, Kim, Yue, Mahler, Krahe, Rodriguez,
  Hodgins, and Matthews}]{taylor2017deep}
Taylor, S.; Kim, T.; Yue, Y.; Mahler, M.; Krahe, J.; Rodriguez, A.~G.; Hodgins,
  J.; and Matthews, I. 2017.
\newblock A deep learning approach for generalized speech animation.
\newblock \emph{TOG} 36(4): 93.

\bibitem[{Thies et~al.(2020)Thies, Elgharib, Tewari, Theobalt, and
  Nie{\ss}ner}]{thies2019neural}
Thies, J.; Elgharib, M.; Tewari, A.; Theobalt, C.; and Nie{\ss}ner, M. 2020.
\newblock Neural Voice Puppetry: Audio-driven Facial Reenactment.
\newblock \emph{ECCV} .

\bibitem[{Thies, Zollh{\"o}fer, and Nie{\ss}ner(2019)}]{thies2019deferred}
Thies, J.; Zollh{\"o}fer, M.; and Nie{\ss}ner, M. 2019.
\newblock Deferred neural rendering: Image synthesis using neural textures.
\newblock \emph{TOG} 38(4): 1--12.

\bibitem[{Thies et~al.(2015)Thies, Zollh{\"o}fer, Nie{\ss}ner, Valgaerts,
  Stamminger, and Theobalt}]{thies2015real}
Thies, J.; Zollh{\"o}fer, M.; Nie{\ss}ner, M.; Valgaerts, L.; Stamminger, M.;
  and Theobalt, C. 2015.
\newblock Real-time expression transfer for facial reenactment.
\newblock \emph{TOG} 34(6): 183--1.

\bibitem[{Thies et~al.(2016)Thies, Zollhofer, Stamminger, Theobalt, and
  Nie{\ss}ner}]{thies2016face2face}
Thies, J.; Zollhofer, M.; Stamminger, M.; Theobalt, C.; and Nie{\ss}ner, M.
  2016.
\newblock Face2face: Real-time face capture and reenactment of rgb videos.
\newblock In \emph{CVPR}, 2387--2395.

\bibitem[{Thies et~al.(2018)Thies, Zollh{\"o}fer, Theobalt, Stamminger, and
  Nie{\ss}ner}]{thies2018headon}
Thies, J.; Zollh{\"o}fer, M.; Theobalt, C.; Stamminger, M.; and Nie{\ss}ner, M.
  2018.
\newblock Headon: Real-time reenactment of human portrait videos.
\newblock \emph{TOG} 37(4): 1--13.

\bibitem[{Vaswani et~al.(2017)Vaswani, Shazeer, Parmar, Uszkoreit, Jones,
  Gomez, Kaiser, and Polosukhin}]{vaswani2017attention}
Vaswani, A.; Shazeer, N.; Parmar, N.; Uszkoreit, J.; Jones, L.; Gomez, A.~N.;
  Kaiser, {\L}.; and Polosukhin, I. 2017.
\newblock Attention is all you need.
\newblock In \emph{NeurIPS}, 5998--6008.

\bibitem[{Vougioukas, Petridis, and Pantic(2019)}]{vougioukas2019end}
Vougioukas, K.; Petridis, S.; and Pantic, M. 2019.
\newblock End-to-End Speech-Driven Realistic Facial Animation with Temporal
  GANs.
\newblock In \emph{CVPRW}, 37--40.

\bibitem[{Wang et~al.(2018)Wang, Liu, Zhu, Tao, Kautz, and
  Catanzaro}]{pix2pixhd}
Wang, T.; Liu, M.; Zhu, J.; Tao, A.; Kautz, J.; and Catanzaro, B. 2018.
\newblock High-resolution image synthesis and semantic manipulation with
  conditional gans.
\newblock In \emph{CVPR}, 8798--8807.

\bibitem[{Wang et~al.(2004)Wang, Bovik, Sheikh, Simoncelli
  et~al.}]{wang2004image}
Wang, Z.; Bovik, A.~C.; Sheikh, H.~R.; Simoncelli, E.~P.; et~al. 2004.
\newblock Image quality assessment: from error visibility to structural
  similarity.
\newblock \emph{TIP} 13(4): 600--612.

\bibitem[{Wiles, Sophia, and Zisserman(2018)}]{wiles2018x2face}
Wiles, O.; Sophia, K.; and Zisserman, A. 2018.
\newblock X2face: A network for controlling face generation using images,
  audio, and pose codes.
\newblock In \emph{ECCV}, 670--686.

\bibitem[{Wu et~al.(2018)Wu, Zhang, Li, Qian, and
  Change~Loy}]{wu2018reenactgan}
Wu, W.; Zhang, Y.; Li, C.; Qian, C.; and Change~Loy, C. 2018.
\newblock Reenactgan: Learning to reenact faces via boundary transfer.
\newblock In \emph{ECCV}, 603--619.

\bibitem[{Yang et~al.(2019)Yang, Dai, Yang, Carbonell, Salakhutdinov, and
  Le}]{yang2019xlnet}
Yang, Z.; Dai, Z.; Yang, Y.; Carbonell, J.; Salakhutdinov, R.~R.; and Le, Q.~V.
  2019.
\newblock Xlnet: Generalized autoregressive pretraining for language
  understanding.
\newblock In \emph{NeurIPS}, 5753--5763.

\bibitem[{Yu et~al.(2018)Yu, Fernando, Ghanem, Porikli, and
  Hartley}]{yu2018face}
Yu, X.; Fernando, B.; Ghanem, B.; Porikli, F.; and Hartley, R. 2018.
\newblock Face super-resolution guided by facial component heatmaps.
\newblock In \emph{ECCV}, 217--233.

\bibitem[{Yu et~al.(2019{\natexlab{a}})Yu, Fernando, Hartley, and
  Porikli}]{yu2019semantic}
Yu, X.; Fernando, B.; Hartley, R.; and Porikli, F. 2019{\natexlab{a}}.
\newblock Semantic face hallucination: Super-resolving very low-resolution face
  images with supplementary attributes.
\newblock \emph{TPAMI} .

\bibitem[{Yu and Porikli(2016)}]{yu2016ultra}
Yu, X.; and Porikli, F. 2016.
\newblock Ultra-resolving face images by discriminative generative networks.
\newblock In \emph{ECCV}, 318--333.

\bibitem[{Yu and Porikli(2017{\natexlab{a}})}]{yu2017face}
Yu, X.; and Porikli, F. 2017{\natexlab{a}}.
\newblock Face hallucination with tiny unaligned images by transformative
  discriminative neural networks.
\newblock In \emph{AAAI}.

\bibitem[{Yu and Porikli(2017{\natexlab{b}})}]{yu2017hallucinating}
Yu, X.; and Porikli, F. 2017{\natexlab{b}}.
\newblock Hallucinating very low-resolution unaligned and noisy face images by
  transformative discriminative autoencoders.
\newblock In \emph{CVPR}, 3760--3768.

\bibitem[{Yu et~al.(2019{\natexlab{b}})Yu, Shiri, Ghanem, and
  Porikli}]{yu2019can}
Yu, X.; Shiri, F.; Ghanem, B.; and Porikli, F. 2019{\natexlab{b}}.
\newblock Can we see more? joint frontalization and hallucination of unaligned
  tiny faces.
\newblock \emph{TPAMI} .

\bibitem[{Zakharov et~al.(2019)Zakharov, Shysheya, Burkov, and
  Lempitsky}]{zakharov2019few}
Zakharov, E.; Shysheya, A.; Burkov, E.; and Lempitsky, V. 2019.
\newblock Few-shot adversarial learning of realistic neural talking head
  models.
\newblock In \emph{ICCV}, 9459--9468.

\bibitem[{Zeng et~al.(2020)Zeng, Pan, Wang, Zhang, and Liu}]{zeng2020realistic}
Zeng, X.; Pan, Y.; Wang, M.; Zhang, J.; and Liu, Y. 2020.
\newblock Realistic Face Reenactment via Self-Supervised Disentangling of
  Identity and Pose.
\newblock In \emph{AAAI}, volume~34, 12757--12764.

\bibitem[{Zhou et~al.(2019)Zhou, Liu, Liu, Luo, and Wang}]{zhou2019talking}
Zhou, H.; Liu, Y.; Liu, Z.; Luo, P.; and Wang, X. 2019.
\newblock Talking face generation by adversarially disentangled audio-visual
  representation.
\newblock In \emph{AAAI}, volume~33, 9299--9306.

\bibitem[{Zhou et~al.(2020)Zhou, Han, Shechtman, Echevarria, Kalogerakis, and
  Li}]{zhou2020makelttalk}
Zhou, Y.; Han, X.; Shechtman, E.; Echevarria, J.; Kalogerakis, E.; and Li, D.
  2020.
\newblock MakeltTalk: speaker-aware talking-head animation.
\newblock \emph{TOG} 39(6): 1--15.

\bibitem[{Zhou et~al.(2018)Zhou, Xu, Landreth, Kalogerakis, Maji, and
  Singh}]{zhou2018visemenet}
Zhou, Y.; Xu, Z.; Landreth, C.; Kalogerakis, E.; Maji, S.; and Singh, K. 2018.
\newblock Visemenet: Audio-driven animator-centric speech animation.
\newblock \emph{TOG} 37(4): 161.

\end{thebibliography}
\end{document}